\documentclass[acmsmall]{acmart}

\AtBeginDocument{%
  }

\setcopyright{acmlicensed}
\copyrightyear{2024}
\acmYear{2024}
\acmDOI{XXXXXXX.XXXXXXX}

\acmVolume{X}
\acmNumber{X}
\acmArticle{X}
\acmMonth{4}

\usepackage{marvosym,wasysym,pifont,utfsym}
\usepackage{newtxtext,newtxmath}
\usepackage{pifont}
\newcommand{\cmark}{\ding{51}}%
\newcommand{\xmark}{\ding{55}}%

\usepackage{amsmath,amsfonts,bm}
\usepackage{algorithm, algorithmicx, algpseudocode}
\algrenewcommand\algorithmicrequire{\textbf{Input:}}
\algrenewcommand\algorithmicensure{\textbf{Output:}}
\usepackage{textcomp}
\usepackage{xcolor}
\usepackage[utf8]{inputenc}
\usepackage{cleveref}
\usepackage[export]{adjustbox}
\usepackage{caption}
\usepackage{subcaption}
\usepackage{booktabs}
\usepackage{multirow}
\usepackage{comment}
\usepackage{url}

\newcommand{\set}[1]{\ensuremath{\mathcal #1}}

\newcommand{\bfx}{{\mathbf x}}
\newcommand{\bfz}{{\mathbf z}}
\newcommand{\bfp}{{\mathbf p}}
\newcommand{\bmeps}{\bm{\epsilon}}
\newcommand{\bfc}{{\mathbf c}}

\newcommand{\bmtheta}{\bm{\theta}}
\newcommand{\bmphi}{\bm{\phi}}
\newcommand{\bmomega}{\bm{\omega}}

\newcommand{\prompt}[1]{``\texttt{#1}''}

\begin{document}

\title{StyleForge: Enhancing Text-to-Image Synthesis for Any Artistic Styles with Dual Binding}

\author{Junseo Park}
\authornote{Both authors contributed equally to this research.}
\email{mki730@dgu.ac.kr}
\author{Beomseok Ko}
\authornotemark[1]
\email{roy7001@dgu.ac.kr}
\author{Hyeryung Jang}
\authornote{Correspondence to: Hyeryung Jang}
\email{hyeryung.jang@dgu.ac.kr}
\affiliation{%
    \institution{Department of Computer Science \& Artificial Intelligence, Dongguk University
    }
    \city{Seoul}
    \country{South Korea}
}



\begin{abstract}
Recent advancements in text-to-image models, such as Stable Diffusion, have showcased their ability to create visual images from natural language prompts. 
However, existing methods like DreamBooth struggle with capturing arbitrary art styles due to the abstract and multifaceted nature of stylistic attributes. 
We introduce Single-StyleForge, a novel approach for personalized text-to-image synthesis across diverse artistic styles. 
Using approximately $15$ to $20$ images of the target style, Single-StyleForge establishes a foundational binding of a unique token identifier with a broad range of attributes of the target style. 
Additionally, auxiliary images are incorporated for dual binding that guides the consistent representation of crucial elements such as people within the target style. 
Furthermore, we present Multi-StyleForge, which enhances image quality and text alignment by binding multiple tokens to partial style attributes. 
Experimental evaluations across six distinct artistic styles demonstrate significant improvements in image quality and perceptual fidelity, as measured by FID, KID, and CLIP scores.
\end{abstract}




\maketitle

\section{Introduction}
\begin{figure*}[h!]
    \centerline{\includegraphics[width=\textwidth]{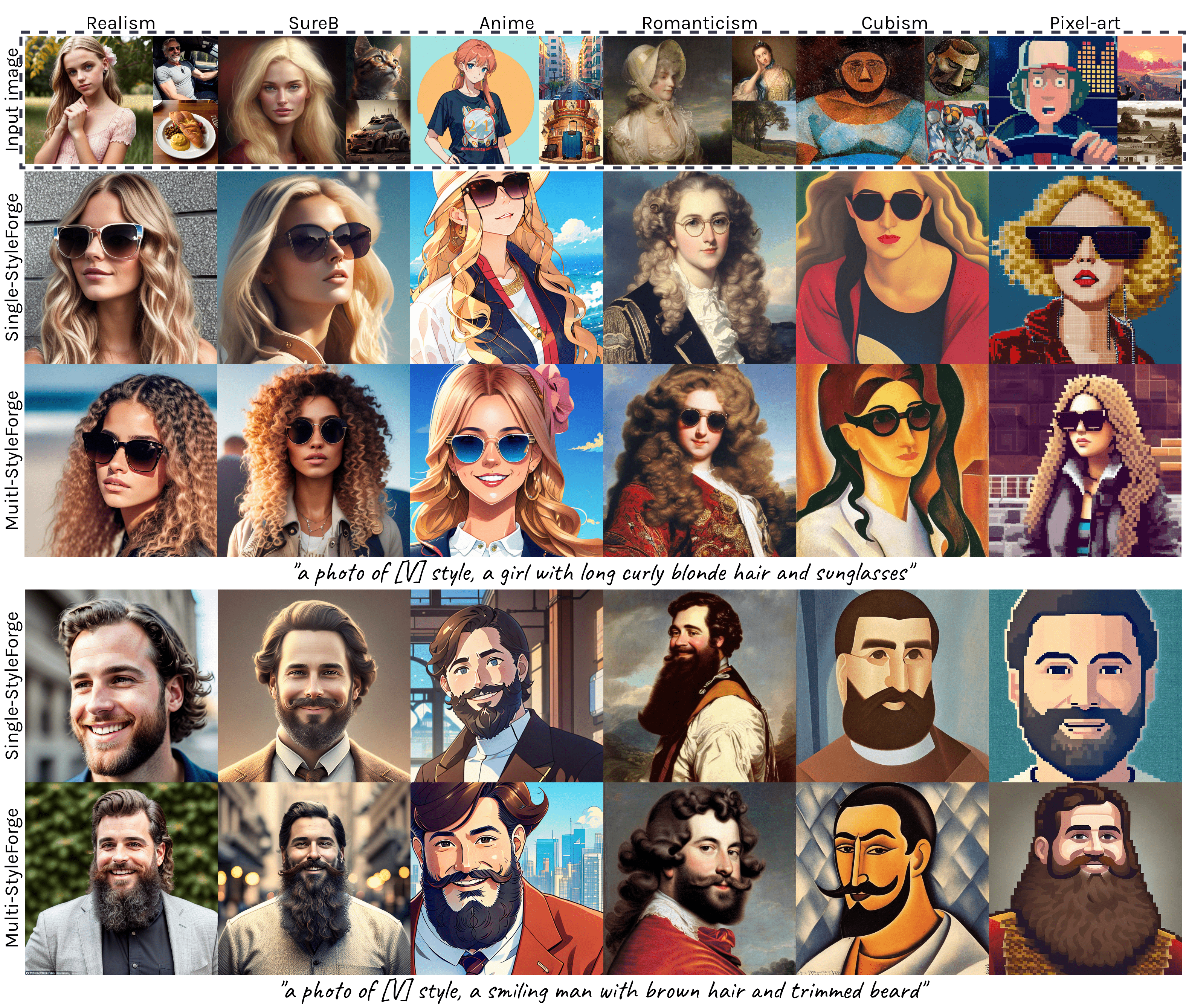}}
    \caption{
    Text-to-image synthesis via Single/Multi-StyleForge personalized in various art styles, from realism to pixel-art. The generated images demonstrate our approach's ability to create aligned and high-fidelity images in each target style (top row) by using a unique token (\prompt{[V] style}) in text prompts. 
    } 
    \label{fig:styleforge_result}
    \vspace{-0.3cm}
\end{figure*}

The field of text-to-image generation has shown remarkable advancements~\cite{dall-e,dall-e2, imagen,parti,stablediffusion} in recent years with the emergence of advanced models such as Stable Diffusion~\cite{stablediffusion}. 
These models facilitate for the creation of intricate visual representations from text inputs, enabling the generation of diverse images based on natural language prompts. 
The significance of these advancements lies not only in their ability to generate images but also in their potential to personalize digital content with user-provided objects and/or styles. 
A notable example is DreamBooth~\cite{dreambooth}, which fine-tunes pre-trained text-to-image models to combine unique text identifiers, or tokens, with a limited set of images representing specific objects. 
This improves the model's adaptability to individual preferences, making it easier to create images of various poses and views of the subject in diverse scenes described by textual prompts. 

Despite significant progress in synthesizing images that mimic the art styles of renowned painters and artistic icons, such as Van Gogh or Pop Art, generating images that encapsulate a broader spectrum of artistic styles remains a challenge. 
The concept of an {\em ``art style''} involves a complex fusion of visual elements, including lines, shapes, textures, and spatial and chromatic relationships, which should be applied in a wide range of landscapes, subjects and people, e.g., \prompt{an Asian girl on a London street in the style of Van Gogh}. These elements are abstract and multifaceted, making it difficult to quantify, classify, and capture them accurately. 
Consequently, in contrast to specific objects, personalizing text-to-image synthesis to generate images aligned with diverse artistic styles requires conveying such abstract and multifaceted nature of visual attributes.
We observe that applying existing personalizing methods, such as DreamBooth ~\cite{dreambooth}, LoRA ~\cite{lora} or Textual Inversion ~\cite{textual-inversion}, to mimic arbitrary art styles is not straightforward enough to yield good performance.

In this paper, we introduce {\bf StyleForge}, a novel approach for personalized text-to-image synthesis across diverse and arbitrary artistic styles. 
Our method leverages pre-trained text-to-image models, e.g., Stable Diffusion, to generate a wide range of images that match specific artistic styles, which we call target styles, guided by text prompts. 
Our contributions are as follows:
\begin{itemize}
    \item We separate stylistic attributes into two main components: backgrounds (i.e., landscape and objects) and persons (i.e., human faces and poses). Representing persons in the target style with high fidelity particularly requires more delicate capture capabilities during training. Existing fine-tuning methods often fail in this aspect, underscoring the need for our specialized approach. 

    \item By utilizing a few reference images showcasing the key characteristics of the target style, along with auxiliary images, Single-StyleForge captures the intricate details of the target style. 
    This involves a dual-binding strategy: establishing a foundational connection between a unique token (e.g., \prompt{[V] style}) and the general features of the target style, and using auxiliary images and prompt (e.g., \prompt{style}) to embed essential aspects (e.g., human features) of the artwork. 

    \item We introduce Multi-StyleForge, which divides the target style into multiple components and maps each to unique identifier during training. Establishing a binding between partial and separate features of the target style and the corresponding prompt, this approach improves the alignment between text prompts and generated images across various styles. 

    \item Through extensive experiments, we evaluate our StyleForge methods with six distinct artistic styles, beyond just well-known examples. We observe substantial performance improvements in both the quality and perceptual fidelity of generated images, as measure by FID, KID and CLIP scores. To summarize, our main result is illustrated in Fig.~\ref{fig:styleforge_result}.

\end{itemize}

\section{related work} \label{sec:related}

\noindent {\bf Text-to-image synthesis.} 
The field of text-to-image synthesis has made significant progress with the emergence of models such as Imagen~\cite{imagen}, DALL-E~\cite{dall-e, dall-e2}, Stable Diffusion (SD)~\cite{stablediffusion}. 
These models use different approaches, including transformer and autoregressive models, as well as diffusion techniques, to create visually appealing images from textual prompts. 
Imagen~\cite{imagen} employs a direct diffusion approach on pixels through a pyramid structure, while Stable Diffusion~\cite{stablediffusion} applies diffusion in the latent space. 
DALLE-E~\cite{dall-e} adopts a transformer-based autoregressive modeling strategy, which is further refined in DALL-E2~\cite{dall-e2} through a two-stage model consisting of a prior and decoder network for diverse image generation. 
Other notable models include Muse~\cite{muse}, which leverages a masked generative image transformer, and Parti~\cite{parti} which combines the ViT-VQGAN~\cite{vq-gan} image tokenizer with an autoregressive model. 

\noindent {\bf Style transfer.} 
Traditional neural style transfer is a technique that alters the visual style of a given image while preserving its content by transferring stylistic characteristics of one image onto another. Early techniques harnessed CNNs or GANs to achieve this; one method ~\cite{vggtransfer} used the feature maps of VGG model to obtain a correlation between style and content images; and StyleGAN~\cite{stylegan} generates high-resolution images by synthesizing styles using a progressive GAN~\cite{pggan} architecture. 
AdAttN~\cite{adaattn} enhances the process by incorporating an attention mechanism, improving upon AdaIN approach~\cite{adain}. 
StyleCLIP~\cite{styleclip} provides intuitive text-based image control for style transfer. 
More recent advancements, such as StyleDiffusion~\cite{style-diffusion}, utilize diffusion models for a more controllable disentanglement of content and style, while PatchMatch~\cite{patchmatch} introduces patch-based technique with whitening and coloring transformations in the diffusion process. 
Inversion-based method~\cite{inversion} allows style transfer via textual descriptions, and DreamStyler~\cite{dreamstyler} leverages language-image models like BLIP-2~\cite{blip} and an image encoder for advanced textual inversion, allowing the binding of style with text in the final output.

\noindent {\bf 
Personalizing/controlling generative models.}  
Recent studies have improved text-to-image synthesis by customizing the model to individual preferences. 
DreamBooth~\cite{dreambooth} introduces personalization by fine-tuning the pre-trained model with a limited set of images of the subject through class-specific prior preservation loss for capturing the semantic details. 
Other works such as CustomDiffusion~\cite{custom-diffusion}, SVDiff~\cite{svdiff}, and HyperDreamBooth~\cite{hyperdreambooth} focus on personalizing multiple subjects simultaneously or in a parameter-efficient manner. 
Another notable approach is textual inversion~\cite{textual-inversion}, which suggests finding embeddings of special text tokens corresponding to specific image sets, whereas ControlNet~\cite{controlnet} introduces a novel structure to control pre-trained models with trainable convolution techniques. 
DreamArtist~\cite{dreamartist} and SpecialistDiffusion~\cite{specialist-diffusion} are recent works focusing on learning positive/negative embeddings and customized data augmentation, respectively, for this purpose. 
Instead of text-to-image diffusion models, StyleDrop~\cite{styledrop} operates on a generative vision transformer Muse~\cite{muse} and can generate diverse visual styles. Finally, we note that advances in parameter-efficient fine-tuning (PEFT) methods like LoRA~\cite{lora} and adapter tuning can contribute to the efficient improvement of generative models for a specific task.  

\section{Preliminaries} \label{sec:prelim}

\noindent {\bf Diffusion models.} 
Diffusion models~\cite{diffusion, 2015-diffusion, ddim} are a class of probabilistic generative models designed to capture and reproduce the distribution of data through an iterative denoising process, starting from random noise. 
These models reverse the diffusion process, which converts a data sample $\bfx_0$ from its unknown distribution $q(\bfx_0)$ into Gaussian noise step by step. The goal is to learn a parameterized model $p_{\bmtheta}(\bfx_0)$ to approximate the distribution $q(\bfx_0)$. 
Diffusion models are interpreted as a sequence of equally weighted denoising autoencoders, $\{ \bmeps_{\bmtheta}(\bfx_t, t)\}_{t = 1 \ldots T}$, trained to predict a denoised variant of their input $\bfx_t$ at each time step $t$ with the objective simplified as:
\begin{align} \label{eq:loss_DM}
    \set{L}_\text{DM} = \mathbb{E}_{\bfx, \bmeps, t} \Big[ \| \bmeps - \bmeps_{\bmtheta}(\bfx_t, t) \|_2^2 \Big],
\end{align}
where $t$ is a time step and $\bmeps \sim \set{N}(0,1)$ is a Gaussian noise. 

On the other hand, the Latent Diffusion Model (LDM) performs denoising operations in the latent space rather than the image space. In LDM, an encoder $\set{E}$ transforms the input image $\bfx$ into a latent code $\bfz = \set{E}(\bfx)$, which is then trained to denoise variably-noised latent code $\bfz_t := \alpha_t \mathcal{E}(\bfx) + \sigma_t \bmeps$ at each step $t$ with the following objective: 
\begin{align} \label{eq:loss_LDM}
    \set{L}_\text{LDM} = \mathbb{E}_{\bfz, \bfc, \bmeps, t} \Big[ \| \bmeps - \bmeps_{\bmtheta}(\bfz_t, t, \bfc) \|_2^2 \Big],
\end{align}
where $\bfc = \Gamma_{\boldsymbol{\phi}}(\bfp)$ is a conditioning vector for some text prompt $\bfp$ obtained by a text encoder $\Gamma_{\boldsymbol{\phi}}$. While training, $\bmeps_{\bmtheta}$ and $\Gamma_{\boldsymbol{\phi}}$ are jointly optimized to minimize the LDM loss \eqref{eq:loss_LDM}, ensuring that the model effectively learns the underlying data distribution for high-quality image synthesis. 


\noindent {\bf DreamBooth.} 
DreamBooth \cite{dreambooth} is a recent method for personalizing pre-trained text-to-image diffusion models, such as LDM, using a few images of a specific object, called instance images. 
By using $3$-$5$ images of the specific object (e.g., my dog) paired with a text prompt (e.g., \prompt{A [V] dog}) containing a unique token identifier (e.g., \prompt{[V]}) that represents the given object (e.g., my dog) and a meta-class name (e.g., \prompt{dog}), DreamBooth fine-tunes a text-to-image model to encode the unique token with the subject while keeping the information of meta-class token. 
To this end, DreamBooth introduces a class-specific prior preservation loss, encouraging the model to maintain semantic knowledge about the meta-class (i.e., dog) and produce diverse instances of it (i.e., various dogs). 

During training, images of the meta-class prior $\bfx^{\text{pr}} = \hat{\bfx}(\bmeps', \bfc^{\text{pr}})$ are sampled from the frozen text-to-image model $\hat{\bfx}$ with initial noise $\bmeps' \sim \mathcal{N}(0, I)$ and conditioning vector $\bfc^{\text{pr}}$ corresponding to the meta-class name (i.e., \prompt{dog}); and the denoising network $\bmeps_{\bmtheta}$ is fine-tuned using a combination of reconstruction loss for both instance images $\bfx$ of the specific object and meta-class prior images $\bfx^\text{pr}$ to successfully denoise latent codes $\bfz_t, \bfz_t^{\text{pr}}$ over the diffusion process $t$. The loss is simply written as follows (see \cite{dreambooth} for the details):
\begin{align} \label{eq:DB-loss-1}
\set{L}_{\text{DB}} = \mathbb{E}_{\bfz, \bfc, \bmeps, \bmeps', t} &\Big[ \| \bmeps - \bmeps_{\bmtheta}(\bfz_t,t , \bfc) \|_2^2 + \lambda \| \bmeps' - \bmeps_{\bmtheta}(\bfz_t^{\text{pr}}, t, \bfc^\text{pr}) \|_2^2 \Big],
\end{align}
where $\lambda$ controls the relative weight of the prior-preservation term, and $\bfz_t$ and $\bfz_t^{\text{pr}} := \alpha_t \mathcal{E}(\bfx^{\text{pr}}) + \sigma_t \bmeps'$ can be obtained from the encoder $\mathcal{E}$ during training. Note that this dual loss \eqref{eq:DB-loss-1} helps the model retain the general characteristics of the meta-class while learning the unique features of the specific object, thereby enhancing its personalization capabilities. 
\section{Method: StyleForge} \label{sec:method}

\begin{figure}[t!]
    \centerline{\includegraphics[width=\columnwidth]{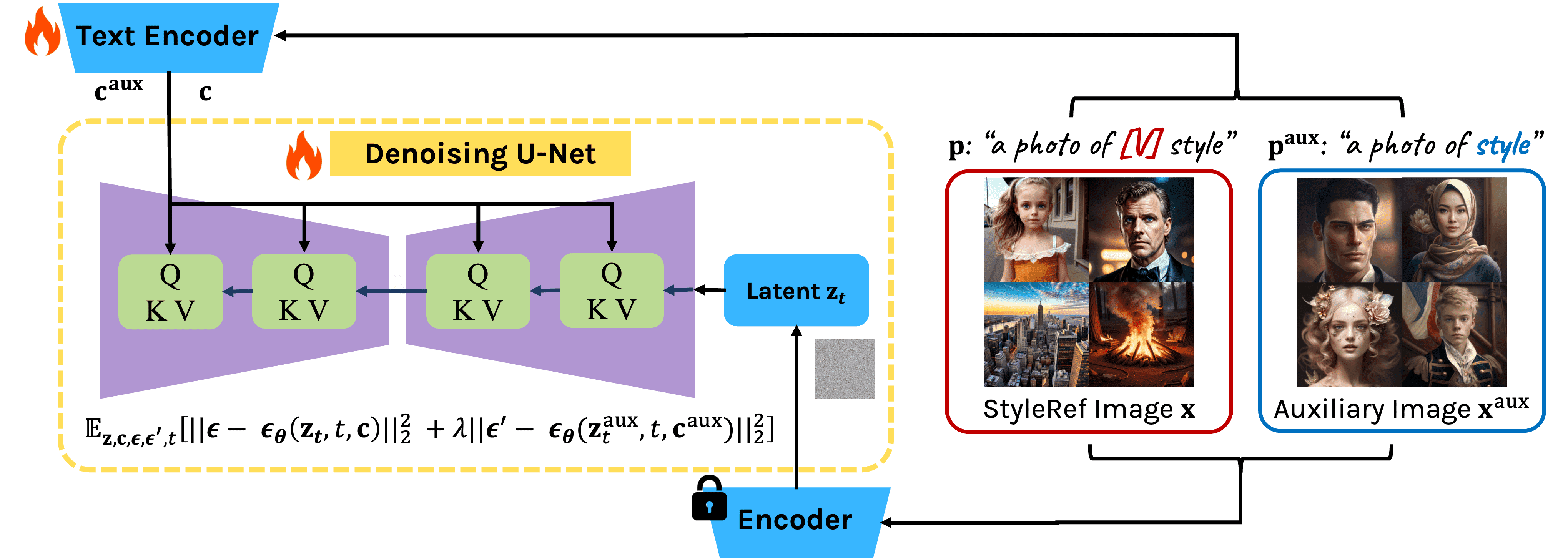}}
    \caption{The architecture of Single-StyleForge. StyleRef images of the target style, paired with text prompt (\prompt{a photo of [V] style}), and Aux images, paired with the prompt (\prompt{a photo of style}), are provided as input images. After fine-tuning, the text-to-image model can generate various images of the target style with the guidance of text prompts. 
    }
    \label{fig:architecture} 
    \vspace{-0.4cm}
\end{figure}

We tackle the challenge of generating high-quality images in various {\em artistic styles}, or simply styles, using text prompts as guidance. While various personalized methods exist, they often face limitations. For example, DreamBooth\cite{dreambooth} does not effectively refine class priority images; energy-efficient fine-tuning techniques such as LoRA\cite{lora}, Textual Inversion\cite{textual-inversion}, and Custom Diffusion\cite{custom-diffusion} only optimize a few parameters of the original model, potentially overlooking specific image details; and techniques that do not train U-Net, like Textual Inversion, may not respond properly to out-of-distribution data. 
These limitations become more apparent with styles that are further from renowned painters, due to the broad and abstract nature of style. 

\subsection{Single-StyleForge: overall architecture} \label{sec:single-architecture}

To address above issues, we first propose {\bf Single-StyleForge}, which reliably generates style variations guided by prompts through comprehensive fine-tuning and auxiliary binding strategies. 
The architecture of Single-StyleForge is based on the framework of DreamBooth\cite{dreambooth}, focusing on synthesizing images of a specific style, which we call {\em target style}, rather than a particular object. 
As illustrated in Fig.~\ref{fig:architecture}, Single-StyleForge fine-tunes pre-trained text-to-image diffusion models $\hat{\bfx}_\theta$ using a few reference images $\bfx$ of the target style, called {\bf StyleRef} images, and a set of auxiliary images $\bfx^\text{aux}$, called {\bf Aux} images. 
The loss function for Single-StyleForge is defined as follows:
\begin{align} \label{eq:SSF-loss}
\set{L}_\text{SSF} = \mathbb{E}_{\bfz, \bfc, \bmeps, \bmeps', t} &\Big[ \| \bmeps - \bmeps_{\bmtheta}(\bfz_t,t , \bfc) \|_2^2 + \lambda \| \bmeps' - \bmeps_{\bmtheta}(\bfz_t^{\text{aux}}, t, \bfc^\text{aux}) \|_2^2 \Big],
\end{align}
where $\bfz_t^{\text{aux}}:= \alpha_t \mathcal{E}(\bfx^{\text{aux}}) + \sigma_t \bmeps'$ and $\bfc^\text{aux}$ are used instead of meta-class prior components in \eqref{eq:DB-loss-1}. In particular, the second term in \eqref{eq:SSF-loss} acts as an auxiliary term that guides the model with information similar to the human perception of the target style. Finally, $\lambda$ controls the strength of the second term. 


For given target style, we use $15$-$20$ StyleRef images $\bfx$ that showcase the key characteristics of the style. These images include a mix of landscapes, objects, and people to provide a comprehensive representation of the target style. The corresponding StyleRef prompts $\bfp$ are crafted to encapsulate the essence of the style using a unique token identifier (e.g., \prompt{[V] style}). 
Unlike the meta-class prior images $\bfx^\text{pr}$ in DreamBooth, the auxiliary images $\bfx^\text{aux}$ are collected to supplement the StyleRef images. These images are chosen to enhance the model's understanding of additional context in the target style, capturing a generic features for challenging elements like human faces and poses. The auxiliary prompts $\bfp^\text{aux}$ use a general token (e.g., \prompt{style}) to ensure they provide broad guidance without introducing bias. 
Finally, the dataset $\set{D}$ for training comprises pairs of StyleRef images with their prompts $(\bfx, \bfp)$ and Aux images with their prompts $(\bfx^\text{aux}, \bfp^\text{aux})$. 

The training process of Single-StyleForge involves the following steps. First, load pre-trained weights for the encoder $\set{E}$, text encoder $\Gamma_{\boldsymbol{\phi}}$, and U-Net $\bmeps_{\bmtheta}$ (line $1$), and sample a pair of StyleRef and Aux images and prompts from the dataset $\set{D}$ (line $3$); then encode the StyleRef and Aux prompts $\bfp, \bfp^\text{aux}$ into latent codes $\bfc, \bfc^\text{aux}$ using the text encoder (line $4$). Next, forward diffusion process is performed to generate noisy latent codes $\bfz_t, \bfz^\text{aux}$ from sampled noises $\bmeps, \bmeps'$ (lines $5$-$7$); and optimize the model by minimizing the loss \eqref{eq:SSF-loss}, taking a gradient descent step to update the model parameters $\bmomega$. The detailed training process is outlined in Algorithm~\ref{alg:SSF}.

\begin{algorithm}[t]
  \caption{Single-StyleForge}
  \label{alg:SSF}
  \begin{algorithmic}[1]
  \Require dataset $\set{D} = \{ (\bfx, \bfp), (\bfx^\text{aux}, \bfp^\text{aux})\}$, encoder \set{E}, text encoder $\Gamma_{\boldsymbol{\phi}}$, U-Net $\bmeps_{\bmtheta}$, hyper-parameters $\{\sigma_t, \alpha_t\}_{t=1,\ldots,T}$ and control parameter $\lambda$   
  \Ensure trained model $\Gamma_{\boldsymbol{\phi}}, \bmeps_{\bmtheta}$ with learnable weights $\bmomega = \{\bmtheta, \bmphi\}$
  \State {\bf Initialize:} load pre-trained weights for $\set{E}, \Gamma_{\bmphi}, \bmeps_{\bmtheta}$
  \Repeat 
    \State sample a data $(\bfx, \bfp), (\bfx^\text{aux}, \bfp^\text{aux}) \sim \set{D}$ 

    \State obtain conditioning vectors $\bfc = \Gamma_{\boldsymbol{\phi}}(\bfp)$,~ $\bfc^\text{aux} = \Gamma_{\boldsymbol{\phi}}(\bfp^\text{aux})$
    \State sample time $t \sim \text{Uniform}(1, \ldots, T)$

    \State sample noise vectors $\bmeps, \bmeps' \sim \set{N}(0, I)$

    \State $\bfz_t := \alpha_t \set{E}(\bfx) + \sigma_t \bmeps$,~ $\bfz_t^\text{aux} := \alpha_t \set{E}(\bfx^\text{aux}) + \sigma_t \bmeps'$ \Comment{forward diffusion processes} 
    \State $\boldsymbol{\omega} \leftarrow \boldsymbol{\omega} -  \text{Optimizer} \Big(\| \bmeps - \bmeps_\theta(\bfz_t,t , \bfc) \|_2^2 + \lambda \| \bmeps' - \bmeps_\theta(\bfz_t^{\text{aux}}, t, \bfc^\text{aux}) \|_2^2 \Big) $ \Comment{\text{gradient descent step}}
  \Until{converged}
  \end{algorithmic}
\end{algorithm}

\subsection{Rationale behind Auxiliary images}
\label{sec:rationale_aux}

The use of auxiliary image $\bfx^\text{aux}$ is crucial in the training process of Single-StyleForge, where its two main roles are discussed here.

\smallskip
\noindent {\bf Aiding in the binding of the target style.} 
While binding a unique token to an object is relatively straightforward as in DreamBooth, 
capturing the diverse features of a target style and combining them with the identifier poses a challenge. 
Due to significant variation in artistic styles, learning features such as vibrant colors, exaggerated facial expressions, and dynamic movements in styles like `anime' becomes difficult with only a few StyleRef images $\bfx$. 
Moreover, we observe that a pre-trained text-to-image model (e.g., Stable Diffusion v1.5) embeds a wide range of characteristics associated with the word \prompt{style}, mostly including fashion styles, fabric patterns, and often art styles, as shown in Fig.~\ref{fig:aux_image}. 
Instead of retaining unnecessary meanings of the word \prompt{style}, we propose utilizing Aux images $\bfx^\text{aux}$ to allow the token \prompt{style} to encapsulate some concepts essential for expressing artwork features. 
As a result, while the StyleRef images $\bfx$ and prompt $\bfp$ (e.g., \prompt{[V] style}) captures overall information about the target style, the Aux images $\bfx^\text{aux}$ and prompt (e.g., \prompt{style}) provides a more detailed information about specific aspects, such as how to represent a person in that style. 
This adjustment redirects the embedding of the word \prompt{style} from unrelated meanings, such as fashion styles, to general artwork styles, alleviating overfitting and thereby enhancing overall learning performance.

\smallskip
\noindent {\bf Improving text-to-image performance.} 
In \cite{dreambooth}, it is demonstrated that using a set of meat-class prior images generated by pre-trained diffusion models a meta-class prompt can enhance personalization capability. 
However, in our case, we found that using a set of images generated by pre-trained diffusion models using a prompt \prompt{style} resulted in unnecessary context such as fashion styles, making it difficult to incorporate essential information during the training process. 
Additionally, we noticed that providing detailed descriptions of a person (e.g., hands, legs, facial features, and full-body poses) remains crucial in qualitative evaluations, while drawing landscapes or animals has less impact. 
Therefore, in Single-StyleForge, Aux images $\bfx^\text{aux}$ primarily consist of portraits and/or people collected from high-resolution images on the Internet, with the aim of enhancing overall text-to-image synthesis performance, particularly for generating high-quality images related to people. 

\smallskip
\noindent {\bf Mitigating language drift.} 
As pointed out in \cite{dreambooth}, personalization of text-to-image models commonly leads to several issues: 
(i) overfitting to a small set of input images (i.e., StyleRef images), resulting in images that are specific to a particular context and subject appearance, such as images with the same background, as well as a lack of alignment between text and image; 
and (ii) language drift, which causes the model to associate the prompt with a limited set of input images and lose diverse meanings of the meta-class name. 
However, when personalizing models to fit a style rather than a specific subject, it is observed that language drift becomes less of a concern, as \prompt{style} is an abstract concept that does not require strict adherence to the meaning and diversity of the word. We expect that if the desired style concept is encoded in the \prompt{style} token, then language drift is not a significant issue.

\subsection{Multi-StyleForge} \label{sec:multi}
While Single-StyleForge focuses on learning a comprehensive representation of a target style, Multi-StyleForge enhances this capability by separating the stylistic attributes into multiple specific components. 
This approach aims to improve the alignment between the text prompts and the generated images, particularly for styles that involve complex compositions of backgrounds and persons. 

Multi-StyleForge builds on the foundation of Single-StyleForge by dividing the components of the target style and mapping each to a unique identifier for training by adopting the method in \cite{custom-diffusion}. 
Since Single-StyleForge mapped StyleRef images to a single prompt (e.g., \prompt{[V] style}), thus during the inference using a prompt without person-related descriptions, it often produces an image that includes a person. 
To address this issue, Multi-StyleForge uses two StyleRef prompts (e.g., \prompt{[V] style, [W] style}), one for persons and another for backgrounds, to train the model more effectively. By separating these elements explicitly, we address the ambiguity that can arise when a single prompt is used to capture both. We also note that it can be extended to separate into multiple components rather than two. 

\smallskip
\noindent {\bf Multi-StyleRef prompts configuration.} 
The StyleRef images consist of two parts: elements of people and backgrounds in target style, following the structure from Single-StyleForge. 
Each component is then associated with its specific prompt (e.g., \prompt{[V] style} for persons and \prompt{[W] style} for backgrounds). The Aux images and prompt $\bfx^\text{aux}, \bfp^\text{aux}$ are kept unified as \prompt{style} to ensure they provide general guidance applicable to both components. 
As a result, Multi-StyleForge trains the model to differentiate stylistic features (people and backgrounds) and obtain separate embeddings.

\begin{algorithm}[t]
  \caption{Multi-StyleForge}
  \label{alg:multi-SB}
  \begin{algorithmic}[1]
  \Require Data $\set{D}_1 = \{ (\bfx, \bfp), (\bfx^\text{aux}, \bfp^\text{aux})\}$, $\set{D}_2 = \{ (\bfx, \bfp), (\bfx^\text{aux}, \bfp^\text{aux})\}$, encoder \set{E}, text encoder $\Gamma_{\boldsymbol{\phi}}$, U-Net $\bmeps_{\bmtheta}$, hyper-parameters $\{\sigma_t, \alpha_t\}_{t=1,\ldots,T}, \lambda$
  \Ensure Trained models $\Gamma_{\boldsymbol{\phi}}, \bmeps_{\bmtheta}$ with learnable weights $\bmomega = \{\bmtheta, \bmphi\}$
  \State {\bf Initialize:} $q = \frac{|\set{D}_1|}{|\set{D}_1| + |\set{D}_2|}$, load pre-trained weights for $\set{E}, \Gamma_{\bmphi}, \bmeps_{\bmtheta}$
  \Repeat 
    \State select a dataset $\set{D} = \set{D}_1$ if $Q \sim \text{Uniform}([0,1]) < q$ else $\set{D}_2$ 
    \State sample a data $(\bfx, \bfp), (\bfx^\text{aux}, \bfp^\text{aux}) \sim \set{D}$ 

    \State $\bfc = \Gamma_{\boldsymbol{\phi}}(\bfp)$, $\bfc^\text{aux} = \Gamma_{\boldsymbol{\phi}}(\bfp^\text{aux})$, sample time $t \sim \text{Uniform}(1, \ldots, T)$

    \State sample a noise $\bmeps, \bmeps' \sim \set{N}(0, I)$

    \State $\bfz_t := \alpha_t \set{E}(\bfx) + \sigma_t \bmeps$, $\bfz_t^\text{aux} := \alpha_t \set{E}(\bfx^\text{aux}) + \sigma_t \bmeps'$ \Comment{forward diffusion processes} 
    \State $\boldsymbol{\omega} \leftarrow \boldsymbol{\omega} -  \text{Optimizer} \Big(\| \bmeps - \bmeps_\theta(\bfz_t,t , \bfc) \|_2^2 + \lambda \| \bmeps' - \bmeps_\theta(\bfz_t^{\text{aux}}, t, \bfc^\text{aux}) \|_2^2 \Big) $ \Comment{gradient descent step}
  \Until{converged}
  \end{algorithmic}
\end{algorithm}

\smallskip
\noindent {\bf Training of Multi-StyleForge.} 
Our Multi-StyleForge is built upon the framework of Custom-Diffusion \cite{custom-diffusion}, which personalizes multiple subjects. The training process of Multi-StyleForge involves simultaneous or parallel learning of the two StyleRef prompts. 
When personalizing each StyleRef prompt sequentially, there is a risk of losing information about previously learned StyleRef prompts in the subsequent process. Therefore, Multi-StyleForge takes simultaneous learning of multiple StyleRef prompts. 
The Multi-StyleForge operation typically involves two text-image pairs $\set{D}_1$ and $\set{D}_2$, which is illustrated in Alg.~\ref{alg:multi-SB}. 
In particular, each component is selected with a probability proportional to the number of data samples (line $3$), then the model is trained using the selected StyleRef data, similar to the Single-StyleForge, to capture the associated stylistic components (lines $4$-$8$). 

Multi-StyleForge improves the alignment between text and images by reducing ambiguity through the use of multiple specific tokens. During inference, these tokens guide the generation process to ensure that images align more accurately with the corresponding components and prompts. For instance, using tokens \prompt{[V]} for persons and \prompt{[W]} for backgrounds helps the model generate images with clear distinctions between these two elements. Experiments conducted with Multi-StyleForge show high fidelity performance and better text-image alignment, which will be discussed in Sec.~\ref{sec:result}. 

\section{Experimental Results} \label{sec:result}

In this section, we assess the performance of Single/Multi-StyleForge in personalizing text-to-image generation across different artistic styles. Through experiments, we evaluate the effectiveness of our methods in producing high-quality images that faithfully adhere to the target styles. 

\subsection{Experimental setup} \label{sec:setup}

We conducted experiments on six common artistic styles: \textsf{realism}, \textsf{SureB}, \textsf{anime}, \textsf{romanticism}, \textsf{cubism}, and \textsf{pixel-art}. The characteristics of each style are summarized as follows:
\begin{itemize}
    \item \textsf{realism} focuses on accurately and detailed representation of subjects. 
    \item \textsf{SureB} combines pragmatic elements with dramatic, exaggerated expressions, creating a fusion of {\em surrealism and baroque} art, thus we call it \textsf{SureB}.
    \item \textsf{anime} refers to a Japanese animation style characterized by vibrant colors, exaggerated facial expressions, and dynamic movement.
    \item \textsf{romanticism} prioritizes emotional expression, imagination, and the sublime, often portraying fantastical and emotional subjects with a focus on rich, dark tones and extensive canvases. 
    \item \textsf{cubism} emphasizes representing visual experiences by depicting objects from multiple angles simultaneously, often in polygonal or fragmented forms.  
    \item \textsf{pixel-art} involves creating images by breaking them down into small square pixels, adjusting their size and arrangement to form the overall image.     
\end{itemize}
Example images for each style are provided in the top row of Fig.~\ref{fig:styleforge_result}, demonstrating various visual attributes of the target styles. 
The StyleRef images for training and evaluation are collected by using pre-trained diffusion models or from the Web. For target styles of \textsf{realism}, \text{SureB} and \textsf{anime}, we used pre-trained diffusion models from Hugging Face~\cite{huggingface} to generate $18,764$ images. For target styles of \textsf{romanticism} and \textsf{cubism}, we collected $3,600$ images from WikiArt~\cite{wikiart}, and we obtained $1,000$ images from Kaggle~\cite{kaggel-pixel-art} for target style of \textsf{pixel-art}. 
In addition, to enhance the ability to generate images of people, we carefully gathered auxiliary images from different auxiliary styles found on the Web. The details of the Aux images are discussed in Sec.~\ref{sec:exp_aux}. 

We evaluate the quality of generated images using various metrics such as FID \cite{FID}, KID \cite{kid}, and CLIP \cite{clip} scores. FID and KID assess the similarity between real and generated images, with lower scores indicating better performance. On the other hand, CLIP score evaluates the alignment between images and text, with higher scores indicating better alignment. 
To create diverse images from the trained model, we utilize $1,562$ prompts from Parti Prompts \cite{parti}, covering $12$ categories, including people, animals, and artwork. Some example prompts include \prompt{the Eiffel Tower}, \prompt{a cat drinking a pint of beer}, and \prompt{a scientist}. 
For each prompt, we produce $12$ images, resulting in a total of $18,744$ images for evaluation. 
In addition, StyleForge trained on the target style generates $6$ images per prompt, resulting in a total of $9,372$ images for each style.



\subsection{Implementation details.} 
\label{sec:implement_details}
\noindent {\bf Ours.} 
As a base diffusion model, we used Stable Diffusion (SD v1.5) that had been pre-trained with realistic images.
We employed the Adam optimizer with a learning rate of $1e^{-6}$, setting inference steps at $30$, with $\lambda$ value of $1$ for the experiments. 
Fine-tuning the pre-trained text-to-image model for six target styles involved minimizing the loss \eqref{eq:SSF-loss} using $20$ StyleRef images and $20$ Aux images. 
All subsequent experiments adhered to training iterations that have achieved the best FID/KID scores, see Fig.~\ref{fig:training_step} in Appendix. 
In Single-StyleForge, we set the StyleRef prompt $\bfp$ as \prompt{a photo of [V] style}, and the Aux prompt $\bfp^{\text{aux}}$ as \prompt{a photo of style}. 
In Multi-StyleForge, StyleRef is divided into two components (i.e., person and background), each paired with \prompt{a photo of [V] style} and \prompt{a photo of [W] style}, respectively. 

\begin{table}[t]
\centering
\caption{Details of baseline methods in terms of fine-tuning method, use of StyleRef and Aux images. Full and partial tuning indicates a tuning of the entire and a subset of the pre-trained model, respectively.}
\resizebox{\linewidth}{!}{
\begin{tabular}{@{}cccccccc@{}}
\toprule
     & DreamBooth & Textual Inversion & LoRA & Custom Diffusion & Single-StyleForge & Multi-StyleForge \\
\midrule
Tuning method & Full & Partial & Partial & Partial & Full & Full \\

StyleRef image& \cmark & \cmark & \cmark & \cmark & \cmark & \cmark \\

Aux image & \cmark & \xmark & \xmark & \cmark & \cmark & \cmark \\
\bottomrule
\end{tabular}}
\label{tab:baseline}
\vspace{-0.3cm}
\end{table}

\smallskip
\noindent {\bf Baseline models.} 
For the baseline models, Textual Inversion~\cite{textual-inversion}, LoRA~\cite{lora}, DreamBooth~\cite{dreambooth}, CustomDiffusion~\cite{custom-diffusion} are trained to achieve the best FID/KID scores using the same text-image pairs for fair comparison. At this time, all CFG scales are set to $7.5$.
Some baseline methods, including Textual Inversion~\cite{textual-inversion} and LoRA~\cite{lora}, do not utilize auxiliary prompts, thus auxiliary images are not used. In the case of DreamBooth~\cite{dreambooth}, auxiliary images were generated by a pre-trained diffusion model using the auxiliary prompt. CustomDiffusion~\cite{custom-diffusion} is used as a baeline of Multi-StyleForge, by using the same text-image pairs, but with auxiliary images being generated as in DreamBooth. See Table~\ref{tab:baseline} for the summary.

\subsection{Analysis of StyleRef images} \label{sec:exp_styleref}

\begin{figure}[t!]
    \centering
    \includegraphics[width=0.65\linewidth]{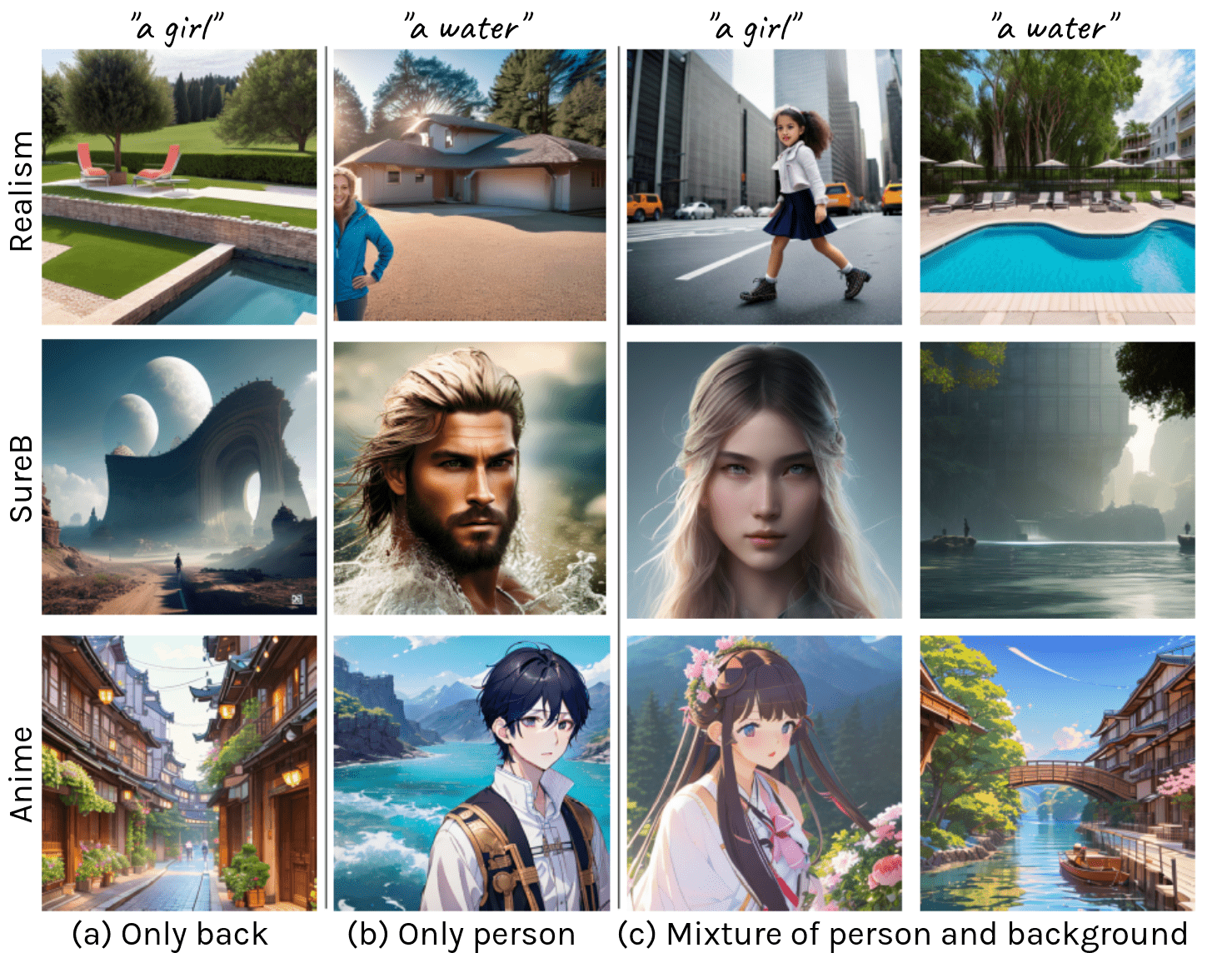}
    \caption{
    Comparison of different compositions of StyleRef images. 
    In (a) and (b), StyleRef images consisting of only background and person, respectively, show that the target style is learned based on biased information, failing to include a girl in (a). The generated images in (c) closely align with the prompts. 
    }
    \label{fig:styleref}
    \vspace{-0.3cm}
\end{figure}

\begin{table}[]
\centering
\caption{FID and KID ($\times 10^3$) scores with different compositions of StyleRef images $\bfx$.}
\resizebox{.9\linewidth}{!}{
\begin{tabular}{@{}cccccccc@{}}
\toprule
\multicolumn{2}{c}{\multirow{2}{*}{StyleRef images}} & \multicolumn{3}{c}{FID score ($\downarrow$)}        & \multicolumn{3}{c}{KID score ($\downarrow$)}             \\ 
\cmidrule(lr){3-5} \cmidrule(lr){6-8}
\multicolumn{2}{c}{}  & \textsf{realism} & \textsf{SureB} & \textsf{anime}  & \textsf{romanticism} & \textsf{cubism} & \textsf{pixel-art} \\ 
\midrule
\multirow{3}{*}{\begin{tabular}[c]{@{}c@{}}  \\  \end{tabular}} &
  only backgrounds &
  $22.804$ &
  $24.598$ &
  $34.629$ & 
  -- &
  -- &
  -- \\
 & only persons   & $21.708$    & $18.812$      & $47.588$ & -- & -- & -- \\
 & mix of backgrounds + persons & $\bm{15.196}$    & $\bm{15.449}$      & $\bm{22.227}$ & $\bm{2.022}$ & $\bm{2.257}$ & $\bm{0.714}$  \\ 
 \bottomrule
\end{tabular}}
\label{tb:styleref}
\vspace{-0.3cm}
\end{table}

We assess the impact of StyleRef images on personalization performance. Encapsulating the target style with numerous StyleRef images may ease customization, it could reduce accessibility, while using only $3$-$5$ images is challenging to capture target style. 
It is important to maintain diversity while personalizing with a limited image set to effectively portray the desired style. Our empirical findings suggest that around $20$ StyleRef images are effective for style personalization. 
To investigate the composition of StyleRef images, we fine-tuned the base diffusion model using $20$ StyleRef images with different combinations, without use of Aux images:
\begin{itemize}
    \item only backgrounds: $20$ landscape images in the target style.
    \item only persons: $20$ portraits and/or people images in the target style. 
    \item mixed backgrounds and persons: a mix of $10$ landscape and $10$ people images.
\end{itemize}
As depicted in Fig.~\ref{fig:styleref}, using only landscape images for StyleRef makes the model misunderstand the target style due to biased information. Similarly, using only people images results in overfitting as the appearance of a person becomes a crucial element, which is shown in Fig.~\ref{fig:styleref}(b). On the other hand, StyleRef images consisting of a mix of landscapes and people effectively capture general features of the target style, synthesizing well-aligned images with the prompts, see Fig.~\ref{fig:styleref}(c). Finally, Table~\ref{tb:styleref} shows a performance improvement ranging from $50.07\%$ to $59.22\%$ in quantitative evaluation when using the mixture StyleRef composition. 


\subsection{Analysis of the Aux images} \label{sec:exp_aux}

\begin{figure}[t]
\centering
\includegraphics[width=.9\textwidth]{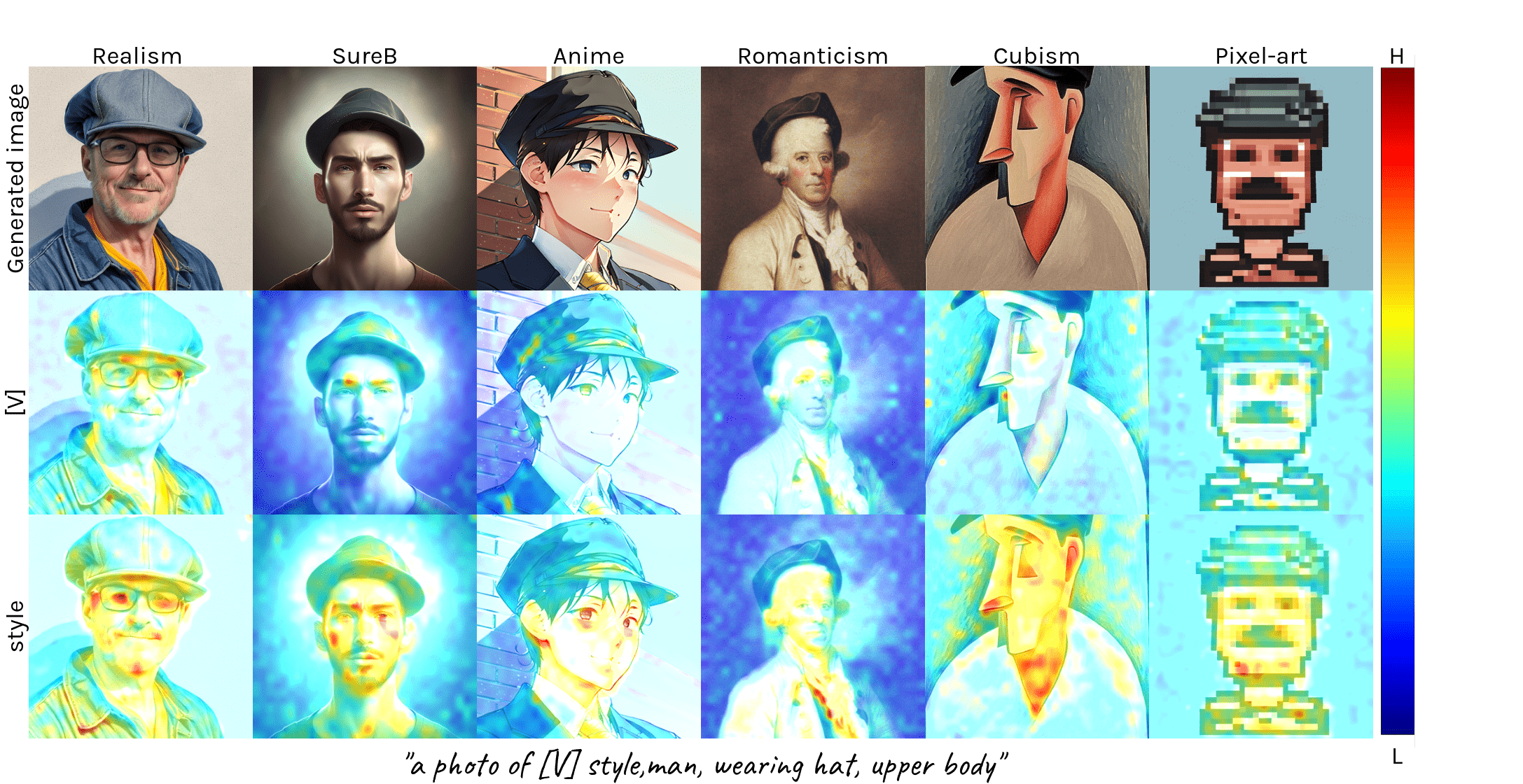}
\caption{Attention maps about \prompt{[V]} and \prompt{style} token in prompt. As we designed, \prompt{[V]} is focusing on a relatively whole area, and \prompt{style} is focusing on people. It was made through edited Prompt-to-Prompt \cite{prompt-to-prompt}.}
\label{fig:04:03}
\vspace{-0.2cm}
\end{figure}

\noindent {\bf Configuration of Aux images.} 
As discussed in Sec.~\ref{sec:rationale_aux}, it is important to properly configure the Aux images properly to improve the performance of Single-StyleForge. When selecting the auxiliary style, we aim to choose a style that is not only similar to the target style but also provides general attributes for creating better images of people. 
With this purpose in mind, we have observed that styles such as \textsf{realism}, \textsf{SureB}, \textsf{anime} and \textsf{romanticism} mainly depict human attributes realistically. In contrast, \textsf{cubism} represents people using polygons and impressionistic shapes, and \textsf{pixel-art} portrays realistic human figures using pixels. 
We have tailored the auxiliary style and images to boost personalization for each target style. For \textsf{realism}, \textsf{SureB}, \text{anime} and \textsf{romanticism} target styles, we collected digital painting images that cover the range from realism to abstraction, while impressionism images were selected for the \textsf{cubism} style; and realism images were chosen for the \textsf{pixel-art} style.

\begin{figure}[t]
\centering
\includegraphics[width=.9\textwidth]{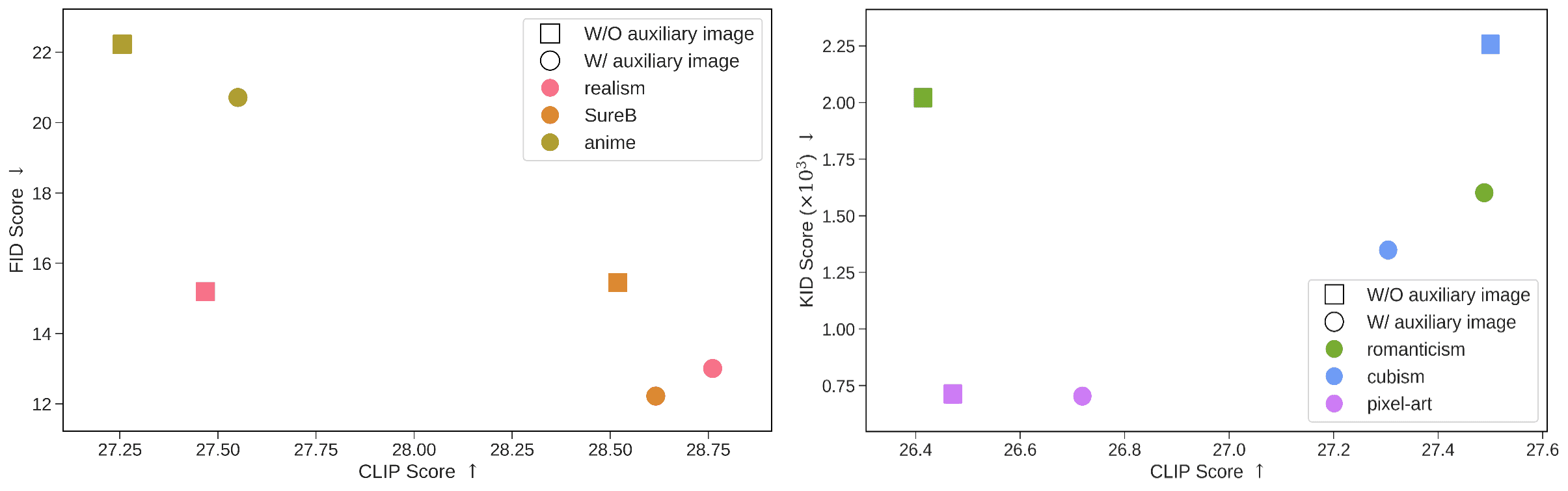}
\caption{Ablation study of Aux images $\bfx^\text{aux}$ for six target styles, displaying FID, KID ($\times 10^3$), and CLIP scores. }
\label{fig:ablation}
\vspace{-0.2cm}
\end{figure} 

\begin{figure}[t]
\centering
\includegraphics[width=.9\textwidth]{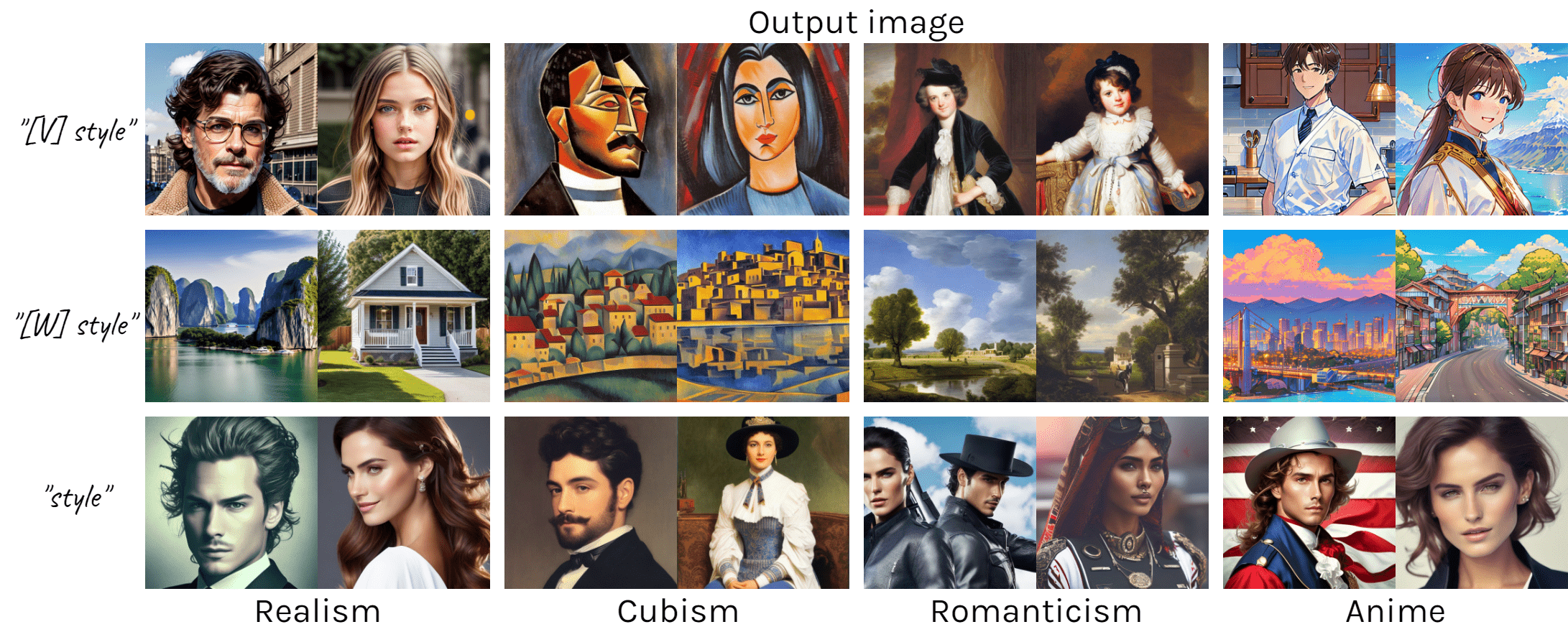}
\caption{Images generated by each token \prompt{[V(person)] style}, \prompt{[W(background)] style} and \prompt{style} from the model trained by Multi-StyleForge. Images in (top) and (middle) include only the person and background, respectively, while images in (bottom) show the person in an auxiliary style.}
    \label{fig:token}
    \vspace{-0.3cm}
\end{figure}

\smallskip
\noindent {\bf Auxiliary binding.}
We designed the Aux images to play a supporting role in the personalization of the target style. Using the Aux images, we transfer the \prompt{style} token from the original fashion style to the artistic style area, facilitating the training process. 
Our main objective is for the \prompt{[V]} token to thoroughly learn the target style, and for the \prompt{style} token to represent people in the auxiliary style as a booster. 
In Fig.~\ref{fig:04:03}, we visualize the attention maps of tokens \prompt{[V]} and \prompt{style} to evaluate the auxiliary binding. 
The attention of \prompt{[V]} is evenly distributed across the images, as the StyleRef images contain both the person and the background, whereas the \prompt{style} token focuses specifically on the person, as the Aux images only contain the person. 
The images generated by each token are displayed in the top and bottom rows of Fig.~\ref{fig:token}. It is observed that \prompt{style} token effectively captures the attributes of the person itself, while \prompt{[V] style} comprehensively produces images of the target style. Finally, the ablation results of using the Aux images are provided in Fig.~\ref{fig:ablation}, showing that adding the Aux images enhances the FID/KID scores across all target styles, indicating a reduction in overfitting, and there is a slight increase in the CLIP score. 

\smallskip
\noindent {\bf Comparison with DreamBooth.}
Table~\ref{tb:02} presents numerical results based on the composition of Aux images, with examples provided in Appendix (Fig.~\ref{fig:aux_image}). Encoding useful information into Aux images enhances performance compared to generating unrefined Aux images through a pre-trained diffusion model proposed by DreamBooth~\cite{dreambooth}. In particular, the performance increases over Style token were $9.02\%$, $14.49\%$, $34.27\%$, $19.86\%$, $63.00\%$, and $16.49\%$. However, composing Aux images with the same style as the target or including dissimilar information (e.g., Human-drawn art) can hinder the model's generalization abilities and lead to overfitting. In summary, while Aux images are not directly linked to the target style, they should complement the style learning process by providing a more comprehensive understanding of visual features and serving as auxiliary bindings for the target style. 

\begin{table}[]
\centering
\caption{Comparison of FID and KID ($\times 10^3$) with different compositions of Aux images $\bfx^\text{aux}$, where we chose the StyleRef composition of backgrounds + persons.
}
\resizebox{.9\linewidth}{!}{
\begin{tabular}{@{}cccccccc@{}}
\toprule
\multicolumn{2}{c}{\multirow{2}{*}{Method}} & \multicolumn{3}{c}{FID score ($\downarrow$)}        & \multicolumn{3}{c}{KID score ($\downarrow$)}             \\ 
\cmidrule(lr){3-5} \cmidrule(lr){6-8}
\multicolumn{2}{c}{}  & \textsf{realism} & \textsf{SureB} & \textsf{anime}  & \textsf{roman} & \textsf{cubism} & \textsf{pixel-art} \\ 
\midrule

\multirow{5}{*}{\begin{tabular}[c]{@{}c@{}} Aux \\ images \end{tabular}} &
  Style token~\cite{dreambooth} &
  $14.297$ &
  $14.293$ &
  $31.518$ &
  $1.999$ & 
  $3.646$ &
  $0.843$ \\
 &
  \begin{tabular}[c]{@{}c@{}}Illustration style\\ token~\cite{dreambooth}\end{tabular} &
  $14.093$ &
  $14.466$ &
  $28.570$ &
  -- &
  -- &
  -- \\ \cmidrule(l){2-8} 
 & Human-drawn art     & $14.263$    & $16.366$      & $22.836$  &
-- &  -- &
  -- \\
 & Target style  & $15.855$    & $13.990$      & $29.450$ & -- &
  -- &
  -- \\
 &
  Single-StyleForge {\bf (ours)} &
  $\bm{13.008}$ &
  $\bm{12.222}$ &
  $\bm{20.718}$ &
  $\bm{1.602}$ & 
  $\bm{1.349}$ &
  $\bm{0.704}$ \\ 
  \bottomrule
\end{tabular}}
\label{tb:02}
\vspace{-0.3cm}
\end{table}

\subsection{Multi-StyleForge: Improved text-image alignment method}

Multi-StyleForge separates styles using specific multiple prompts to better distinguish between images generated from different text conditions. This approach clarifies distinctions between people and backgrounds, thereby improving text alignment. During inference, the prompt \prompt{[V]} is used if the text involves a person, \prompt{[W]} for the background, and \prompt{[V], [W]} if both are relevant. 
Fig.~\ref{fig:token} evaluates the effectiveness of each prompt component in separating people and backgrounds. Specifically, the prompt \prompt{[V(person) style]} is observed to generate images of a person in the target style, while the prompt \prompt{[W(background) style]} creates images of various backgrounds in the target style. 
Fig.~\ref{fig:styleforge_result},~\ref{fig:comparison_total},~\ref{fig:comparison1} and ~\ref{fig:comparison2} compare the image outputs using Multi-StyleForge and Single-StyleForge with various prompts, highlighting their effects on text alignment. By examining the texture of six target styles, detailed background representation, and the shapes of people and details of their faces in the generated images in these figures, it becomes evident that Multi-StyleForge excels at rendering these elements based on text descriptions. 
For instance, when evaluating the presence of ``sunglasses'' in target style \textsf{romanticism} in the top row of Fig.~\ref{fig:styleforge_result}, such details are observed only in the images generated by Multi-StyleForge. 
As shown in Table.~\ref{tb:comparison}, Multi-StyleForge significantly improves text alignment compared to other baselines, as evidenced by higher CLIP scores for all target styles except \textsf{anime}, which shows the second-highest score with Multi-StyleForge.


\subsection{Comparison} \label{sec:comparison}

\begin{table}[t]
\centering
\caption{Quantitative comparisons with FID, KID ($\times 10^3$), and CLIP scores. The table presents FID scores for realism, SureB, and anime styles, along with KID scores for romanticism, cubism, and pixel-art styles, and CLIP scores for all styles. \textbf{Bold} and \underline{underline} denote the best and the second best result, respectively.}  
\resizebox{\textwidth}{!}{%
\begin{tabular}{ccccccccccccc}
\toprule
\multirow{2}{*}{Method} & \multicolumn{3}{c}{FID score ($\downarrow$)}        & \multicolumn{3}{c}{KID score ($\downarrow$)} & \multicolumn{6}{c}{CLIP score ($\uparrow$)}           \\ \cmidrule(rl){2-4} \cmidrule(rl){5-7} \cmidrule(rl){8-13}
                  & \textsf{realism} & \textsf{SureB} & \textsf{anime}  & \textsf{roman} & \textsf{cubism} & \textsf{pixel-art} & \textsf{realism} & \textsf{SureB} & \textsf{anime}  & \textsf{roman} & \textsf{cubism} & \textsf{pixel-art}  \\ 
  \midrule
DreamBooth~\cite{dreambooth}        & $14.093$      & $14.293$    & $28.570$ & $1.999$    & $3.646$      & $\underline{0.843}$ & $28.226$ & $29.020$ & $28.551$ & $27.420$ & $27.818$ & $26.175$  \\
Textual Inversion~\cite{textual-inversion}  & $17.048$  & $22.797$    & $41.654$ & $6.113$         & $4.783$           & $2.330$  & $28.227$ & $27.063$ & $26.284$  & $25.482$ & $22.984$ & $26.497$  \\
LoRA~\cite{lora}             & $\underline{13.218}$         & $16.247$    & $24.560$ & $8.664$        & $13.183$          & $2.641$ & $\underline{28.926}$ & $\underline{29.406}$ & $\bm{29.015}$ & $\underline{29.074}$ & $\underline{28.188}$ & $\underline{29.534}$      \\
Custom Diffusion~\cite{custom-diffusion}   & $21.906$         & $20.227$      & $35.948$    & $7.544$         & $6.680$           & $2.481$    &  $28.253$ &   $29.012$   &   $28.246$   & $26.452$ & $27.395$ &$25.424$ \\
Single-StyleForge {\bf (ours)}        & $\bm{13.008}$ & $\bm{12.222}$ & $\bm{20.718}$ & $\bm{1.602}$ & $\bm{1.349}$ & $\bm{0.704}$ & $28.761$ & $28.616$ & $27.551$  & $27.488$ & $27.304$ & $26.719$  \\
Multi-StyleForge {\bf (ours)}  & $13.480$ & $\underline{12.764}$ & $\underline{20.880}$ & $\underline{1.912}$ & $\underline{1.820}$ & $1.216$  & $\bm{31.215}$ & $\bm{32.082}$ & $\underline{28.662}$ & $\bm{31.243}$ & $\bm{30.891}$ & $\bm{29.852}$  \\ 
\bottomrule
\end{tabular}%
}
\label{tb:comparison}
\vspace{-0.25cm}
\end{table}

Finally, we compare our Single and Multi-StyleForge methods with existing baseline methods. First, quantitative comparisons in terms of FID/KID and CLIP scores are provided in Table~\ref{tb:comparison}. 
We found that the DreamBooth~\cite{dreambooth} generally performs well but its effectiveness decreases for styles such as \textsf{anime}, \textsf{cubism} and \textsf{pixel-art}. 
For Textual Inversion~\cite{textual-inversion}, as the target style becomes less realistic, both FID/KID and CLIP scores decrease significantly. 
Across all target styles, Single-StyleForge demonstrated superior performance in terms of FID/KID scores, followed by Multi-StyleForge. 
Regarding CLIP scores, Multi-StyleForge achieved the best performance by improving text-image alignment. 
Compared to Custom Diffusion~\cite{custom-diffusion} and LoRA~\cite{lora}, which use the concept of personalizing multi-subjects and/or parameter-efficient fine-tuning, it is evident that our Single/Multi-StyleForge methods using full-tuning perform better in learning the ambiguous subject ``style''. 

\begin{figure}[t!]
    \hspace*{-2cm}
    \centering
    \includegraphics[width=\textwidth]{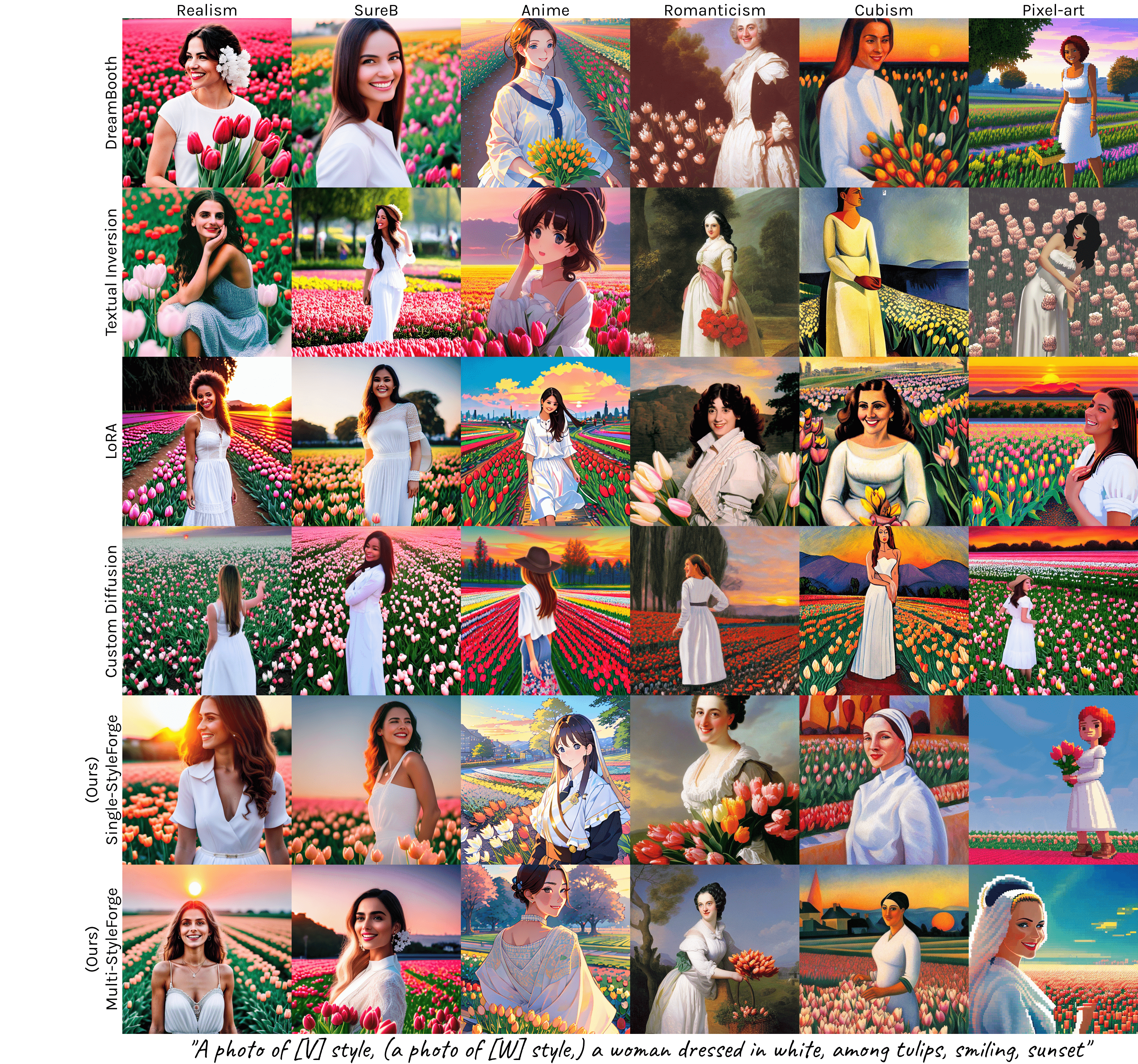}
    \caption{Comparison of our methods to existing personalization techniques. The images are guided by prompts related to humans and backgrounds.}
    \label{fig:comparison_total}
    \vspace{-0.25cm}
\end{figure}

A qualitative comparison with baselines is illustrated in Fig.~\ref{fig:comparison_total}. Some methods often involve a trade-off between reflecting the target artistic style and aligning images with text.
In the case of \textsf{pixel-art}, Textual Inversion and LoRA struggle to fully capture the target style. Furthermore, Textual Inversion (in \textsf{cubism}) and Custom Diffusion fail to depict details like \prompt{smiling}, with Custom Diffusion often showing a rear view instead.
In contrast, our methods faithfully capture both the target style and the text in the generated images. More images generated using other text prompts are provided in the Appendix (see Fig.~\ref{fig:comparison1} and ~\ref{fig:comparison2}). 

\vspace{3cm}
\section{Conclusion} \label{sec:concl}

We introduced an innovative approach for personalized image generation that extends to abstract concepts using a pre-trained text-to-image diffusion model. Single-StyleForge transforms the concept of auxiliary images to introduce dual binding, while Multi-StyleForge enhances this by improving text-image alignment. Extensive experimental results demonstrate that Single/Multi-StyleForge achieve superior text-image alignment with fewer reference images and training steps, achieving remarkable style consistency and high fidelity in text-to-image synthesis.


\section*{Acknowledgement}
This work was supported by the Ministry of Science and ICT, Korea, under the ITRC(Information Technology Research Center) support program(IITP-2023-2020-0-01789), and the Artificial Intelligence Convergence Innovation Human Resources Development (IITP-2023-RS-2023-00254592) supervised by the IITP(Institute for Information \& Communications Technology Planning \& Evaluation).


\begin{thebibliography}{40}
%

\ifx \showCODEN    \undefined \def \showCODEN     #1{\unskip}     \fi
\ifx \showDOI      \undefined \def \showDOI       #1{#1}\fi
\ifx \showISBNx    \undefined \def \showISBNx     #1{\unskip}     \fi
\ifx \showISBNxiii \undefined \def \showISBNxiii  #1{\unskip}     \fi
\ifx \showISSN     \undefined \def \showISSN      #1{\unskip}     \fi
\ifx \showLCCN     \undefined \def \showLCCN      #1{\unskip}     \fi
\ifx \shownote     \undefined \def \shownote      #1{#1}          \fi
\ifx \showarticletitle \undefined \def \showarticletitle #1{#1}   \fi
\ifx \showURL      \undefined \def \showURL       {\relax}        \fi
\providecommand\bibfield[2]{#2}
\providecommand\bibinfo[2]{#2}
\providecommand\natexlab[1]{#1}
\providecommand\showeprint[2][]{arXiv:#2}

\bibitem[hug({[n.\,d.]})]%
        {huggingface}
 \bibinfo{year}{[n.\,d.]}\natexlab{}.
\newblock \bibinfo{title}{Hugging Face}.
\newblock
\newblock
\newblock
\shownote{\url/https://huggingface.co}.


\bibitem[kag({[n.\,d.]})]%
        {kaggel-pixel-art}
 \bibinfo{year}{[n.\,d.]}\natexlab{}.
\newblock \bibinfo{title}{pixel-art}.
\newblock
\newblock
\newblock
\shownote{\url/https://www.kaggle.com/datasets}.


\bibitem[wik({[n.\,d.]})]%
        {wikiart}
 \bibinfo{year}{[n.\,d.]}\natexlab{}.
\newblock \bibinfo{title}{WikiArt}.
\newblock
\newblock
\newblock
\shownote{\url/https://www.wikiart.org/}.


\bibitem[Ahn et~al\mbox{.}(2023)]%
        {dreamstyler}
\bibfield{author}{\bibinfo{person}{Namhyuk Ahn}, \bibinfo{person}{Junsoo Lee}, \bibinfo{person}{Chunggi Lee}, \bibinfo{person}{Kunhee Kim}, \bibinfo{person}{Daesik Kim}, \bibinfo{person}{Seung-Hun Nam}, {and} \bibinfo{person}{Kibeom Hong}.} \bibinfo{year}{2023}\natexlab{}.
\newblock \showarticletitle{DreamStyler: Paint by Style Inversion with Text-to-Image Diffusion Models}.
\newblock \bibinfo{journal}{\emph{arXiv preprint arXiv:2309.06933}} (\bibinfo{year}{2023}).
\newblock


\bibitem[BiÅ„kowski et~al\mbox{.}(2021)]%
        {kid}
\bibfield{author}{\bibinfo{person}{MikoÅ‚aj BiÅ„kowski}, \bibinfo{person}{Danica~J. Sutherland}, \bibinfo{person}{Michael Arbel}, {and} \bibinfo{person}{Arthur Gretton}.} \bibinfo{year}{2021}\natexlab{}.
\newblock \bibinfo{title}{Demystifying MMD GANs}.
\newblock
\newblock
\showeprint[arxiv]{1801.01401}~[stat.ML]


\bibitem[Chang et~al\mbox{.}(2023)]%
        {muse}
\bibfield{author}{\bibinfo{person}{Huiwen Chang}, \bibinfo{person}{Han Zhang}, \bibinfo{person}{Jarred Barber}, \bibinfo{person}{AJ Maschinot}, \bibinfo{person}{Jose Lezama}, \bibinfo{person}{Lu Jiang}, \bibinfo{person}{Ming-Hsuan Yang}, \bibinfo{person}{Kevin Murphy}, \bibinfo{person}{William~T. Freeman}, \bibinfo{person}{Michael Rubinstein}, \bibinfo{person}{Yuanzhen Li}, {and} \bibinfo{person}{Dilip Krishnan}.} \bibinfo{year}{2023}\natexlab{}.
\newblock \bibinfo{title}{Muse: Text-To-Image Generation via Masked Generative Transformers}.
\newblock
\newblock
\showeprint[arxiv]{2301.00704}~[cs.CV]


\bibitem[Dong et~al\mbox{.}(2023)]%
        {dreamartist}
\bibfield{author}{\bibinfo{person}{Ziyi Dong}, \bibinfo{person}{Pengxu Wei}, {and} \bibinfo{person}{Liang Lin}.} \bibinfo{year}{2023}\natexlab{}.
\newblock \bibinfo{title}{DreamArtist: Towards Controllable One-Shot Text-to-Image Generation via Positive-Negative Prompt-Tuning}.
\newblock
\newblock
\showeprint[arxiv]{2211.11337}~[cs.CV]


\bibitem[Gal et~al\mbox{.}(2022)]%
        {textual-inversion}
\bibfield{author}{\bibinfo{person}{Rinon Gal}, \bibinfo{person}{Yuval Alaluf}, \bibinfo{person}{Yuval Atzmon}, \bibinfo{person}{Or Patashnik}, \bibinfo{person}{Amit~H Bermano}, \bibinfo{person}{Gal Chechik}, {and} \bibinfo{person}{Daniel Cohen-Or}.} \bibinfo{year}{2022}\natexlab{}.
\newblock \showarticletitle{An image is worth one word: Personalizing text-to-image generation using textual inversion}.
\newblock \bibinfo{journal}{\emph{arXiv preprint arXiv:2208.01618}} (\bibinfo{year}{2022}).
\newblock


\bibitem[Gatys et~al\mbox{.}(2016)]%
        {vggtransfer}
\bibfield{author}{\bibinfo{person}{Leon~A Gatys}, \bibinfo{person}{Alexander~S Ecker}, {and} \bibinfo{person}{Matthias Bethge}.} \bibinfo{year}{2016}\natexlab{}.
\newblock \showarticletitle{Image style transfer using convolutional neural networks}. In \bibinfo{booktitle}{\emph{Proceedings of the IEEE conference on computer vision and pattern recognition}}. \bibinfo{pages}{2414--2423}.
\newblock


\bibitem[Hamazaspyan and Navasardyan(2023)]%
        {patchmatch}
\bibfield{author}{\bibinfo{person}{Mark Hamazaspyan} {and} \bibinfo{person}{Shant Navasardyan}.} \bibinfo{year}{2023}\natexlab{}.
\newblock \showarticletitle{Diffusion-Enhanced PatchMatch: A Framework for Arbitrary Style Transfer With Diffusion Models}. In \bibinfo{booktitle}{\emph{Proceedings of the IEEE/CVF Conference on Computer Vision and Pattern Recognition}}. \bibinfo{pages}{797--805}.
\newblock


\bibitem[Han et~al\mbox{.}(2023)]%
        {svdiff}
\bibfield{author}{\bibinfo{person}{Ligong Han}, \bibinfo{person}{Yinxiao Li}, \bibinfo{person}{Han Zhang}, \bibinfo{person}{Peyman Milanfar}, \bibinfo{person}{Dimitris Metaxas}, {and} \bibinfo{person}{Feng Yang}.} \bibinfo{year}{2023}\natexlab{}.
\newblock \bibinfo{title}{SVDiff: Compact Parameter Space for Diffusion Fine-Tuning}.
\newblock
\newblock
\showeprint[arxiv]{2303.11305}~[cs.CV]


\bibitem[Hertz et~al\mbox{.}(2022)]%
        {prompt-to-prompt}
\bibfield{author}{\bibinfo{person}{Amir Hertz}, \bibinfo{person}{Ron Mokady}, \bibinfo{person}{Jay Tenenbaum}, \bibinfo{person}{Kfir Aberman}, \bibinfo{person}{Yael Pritch}, {and} \bibinfo{person}{Daniel Cohen-Or}.} \bibinfo{year}{2022}\natexlab{}.
\newblock \bibinfo{title}{Prompt-to-Prompt Image Editing with Cross Attention Control}.
\newblock
\newblock
\showeprint[arxiv]{2208.01626}~[cs.CV]


\bibitem[Hessel et~al\mbox{.}(2021)]%
        {clip-score}
\bibfield{author}{\bibinfo{person}{Jack Hessel}, \bibinfo{person}{Ari Holtzman}, \bibinfo{person}{Maxwell Forbes}, \bibinfo{person}{Ronan~Le Bras}, {and} \bibinfo{person}{Yejin Choi}.} \bibinfo{year}{2021}\natexlab{}.
\newblock \showarticletitle{Clipscore: A reference-free evaluation metric for image captioning}.
\newblock \bibinfo{journal}{\emph{arXiv preprint arXiv:2104.08718}} (\bibinfo{year}{2021}).
\newblock


\bibitem[Heusel et~al\mbox{.}(2017)]%
        {FID}
\bibfield{author}{\bibinfo{person}{Martin Heusel}, \bibinfo{person}{Hubert Ramsauer}, \bibinfo{person}{Thomas Unterthiner}, \bibinfo{person}{Bernhard Nessler}, {and} \bibinfo{person}{Sepp Hochreiter}.} \bibinfo{year}{2017}\natexlab{}.
\newblock \showarticletitle{Gans trained by a two time-scale update rule converge to a local nash equilibrium}. In \bibinfo{booktitle}{\emph{Advances in Neural Information Processing Systems}}, Vol.~\bibinfo{volume}{30}.
\newblock


\bibitem[Ho et~al\mbox{.}(2020)]%
        {diffusion}
\bibfield{author}{\bibinfo{person}{Jonathan Ho}, \bibinfo{person}{Ajay Jain}, {and} \bibinfo{person}{Pieter Abbeel}.} \bibinfo{year}{2020}\natexlab{}.
\newblock \showarticletitle{Denoising diffusion probabilistic models}.
\newblock \bibinfo{journal}{\emph{Advances in Neural Information Processing Systems}}  \bibinfo{volume}{33} (\bibinfo{year}{2020}), \bibinfo{pages}{6840--6851}.
\newblock


\bibitem[Hu et~al\mbox{.}(2021)]%
        {lora}
\bibfield{author}{\bibinfo{person}{Edward~J Hu}, \bibinfo{person}{Yelong Shen}, \bibinfo{person}{Phillip Wallis}, \bibinfo{person}{Zeyuan Allen-Zhu}, \bibinfo{person}{Yuanzhi Li}, \bibinfo{person}{Shean Wang}, \bibinfo{person}{Lu Wang}, {and} \bibinfo{person}{Weizhu Chen}.} \bibinfo{year}{2021}\natexlab{}.
\newblock \showarticletitle{Lora: Low-rank adaptation of large language models}.
\newblock \bibinfo{journal}{\emph{arXiv preprint arXiv:2106.09685}} (\bibinfo{year}{2021}).
\newblock


\bibitem[Huang and Belongie(2017)]%
        {adain}
\bibfield{author}{\bibinfo{person}{Xun Huang} {and} \bibinfo{person}{Serge Belongie}.} \bibinfo{year}{2017}\natexlab{}.
\newblock \showarticletitle{Arbitrary style transfer in real-time with adaptive instance normalization}. In \bibinfo{booktitle}{\emph{Proceedings of the IEEE international conference on computer vision}}. \bibinfo{pages}{1501--1510}.
\newblock


\bibitem[Karras et~al\mbox{.}(2017)]%
        {pggan}
\bibfield{author}{\bibinfo{person}{Tero Karras}, \bibinfo{person}{Timo Aila}, \bibinfo{person}{Samuli Laine}, {and} \bibinfo{person}{Jaakko Lehtinen}.} \bibinfo{year}{2017}\natexlab{}.
\newblock \showarticletitle{Progressive growing of gans for improved quality, stability, and variation}.
\newblock \bibinfo{journal}{\emph{arXiv preprint arXiv:1710.10196}} (\bibinfo{year}{2017}).
\newblock


\bibitem[Karras et~al\mbox{.}(2019)]%
        {stylegan}
\bibfield{author}{\bibinfo{person}{Tero Karras}, \bibinfo{person}{Samuli Laine}, {and} \bibinfo{person}{Timo Aila}.} \bibinfo{year}{2019}\natexlab{}.
\newblock \showarticletitle{A style-based generator architecture for generative adversarial networks}. In \bibinfo{booktitle}{\emph{Proceedings of the IEEE/CVF conference on computer vision and pattern recognition}}. \bibinfo{pages}{4401--4410}.
\newblock


\bibitem[Kumari et~al\mbox{.}(2023)]%
        {custom-diffusion}
\bibfield{author}{\bibinfo{person}{Nupur Kumari}, \bibinfo{person}{Bingliang Zhang}, \bibinfo{person}{Richard Zhang}, \bibinfo{person}{Eli Shechtman}, {and} \bibinfo{person}{Jun-Yan Zhu}.} \bibinfo{year}{2023}\natexlab{}.
\newblock \showarticletitle{Multi-concept customization of text-to-image diffusion}. In \bibinfo{booktitle}{\emph{Proc. of the IEEE/CVF Conference on Computer Vision and Pattern Recognition}}. \bibinfo{pages}{1931--1941}.
\newblock


\bibitem[Li et~al\mbox{.}(2023)]%
        {blip}
\bibfield{author}{\bibinfo{person}{Junnan Li}, \bibinfo{person}{Dongxu Li}, \bibinfo{person}{Silvio Savarese}, {and} \bibinfo{person}{Steven Hoi}.} \bibinfo{year}{2023}\natexlab{}.
\newblock \showarticletitle{Blip-2: Bootstrapping language-image pre-training with frozen image encoders and large language models}.
\newblock \bibinfo{journal}{\emph{arXiv preprint arXiv:2301.12597}} (\bibinfo{year}{2023}).
\newblock


\bibitem[Liu et~al\mbox{.}(2021)]%
        {adaattn}
\bibfield{author}{\bibinfo{person}{Songhua Liu}, \bibinfo{person}{Tianwei Lin}, \bibinfo{person}{Dongliang He}, \bibinfo{person}{Fu Li}, \bibinfo{person}{Meiling Wang}, \bibinfo{person}{Xin Li}, \bibinfo{person}{Zhengxing Sun}, \bibinfo{person}{Qian Li}, {and} \bibinfo{person}{Errui Ding}.} \bibinfo{year}{2021}\natexlab{}.
\newblock \showarticletitle{Adaattn: Revisit attention mechanism in arbitrary neural style transfer}. In \bibinfo{booktitle}{\emph{Proceedings of the IEEE/CVF international conference on computer vision}}. \bibinfo{pages}{6649--6658}.
\newblock


\bibitem[Lu et~al\mbox{.}(2023)]%
        {specialist-diffusion}
\bibfield{author}{\bibinfo{person}{Haoming Lu}, \bibinfo{person}{Hazarapet Tunanyan}, \bibinfo{person}{Kai Wang}, \bibinfo{person}{Shant Navasardyan}, \bibinfo{person}{Zhangyang Wang}, {and} \bibinfo{person}{Humphrey Shi}.} \bibinfo{year}{2023}\natexlab{}.
\newblock \showarticletitle{Specialist Diffusion: Plug-and-Play Sample-Efficient Fine-Tuning of Text-to-Image Diffusion Models To Learn Any Unseen Style}. In \bibinfo{booktitle}{\emph{Proceedings of the IEEE/CVF Conference on Computer Vision and Pattern Recognition}}. \bibinfo{pages}{14267--14276}.
\newblock


\bibitem[Meng et~al\mbox{.}(2021)]%
        {sdedit}
\bibfield{author}{\bibinfo{person}{Chenlin Meng}, \bibinfo{person}{Yutong He}, \bibinfo{person}{Yang Song}, \bibinfo{person}{Jiaming Song}, \bibinfo{person}{Jiajun Wu}, \bibinfo{person}{Jun-Yan Zhu}, {and} \bibinfo{person}{Stefano Ermon}.} \bibinfo{year}{2021}\natexlab{}.
\newblock \showarticletitle{Sdedit: Guided image synthesis and editing with stochastic differential equations}.
\newblock \bibinfo{journal}{\emph{arXiv preprint arXiv:2108.01073}} (\bibinfo{year}{2021}).
\newblock


\bibitem[Patashnik et~al\mbox{.}(2021)]%
        {styleclip}
\bibfield{author}{\bibinfo{person}{Or Patashnik}, \bibinfo{person}{Zongze Wu}, \bibinfo{person}{Eli Shechtman}, \bibinfo{person}{Daniel Cohen-Or}, {and} \bibinfo{person}{Dani Lischinski}.} \bibinfo{year}{2021}\natexlab{}.
\newblock \showarticletitle{Styleclip: Text-driven manipulation of stylegan imagery}. In \bibinfo{booktitle}{\emph{Proceedings of the IEEE/CVF International Conference on Computer Vision}}. \bibinfo{pages}{2085--2094}.
\newblock


\bibitem[Radford et~al\mbox{.}(2021)]%
        {clip}
\bibfield{author}{\bibinfo{person}{Alec Radford}, \bibinfo{person}{Jong~Wook Kim}, \bibinfo{person}{Chris Hallacy}, \bibinfo{person}{Aditya Ramesh}, \bibinfo{person}{Gabriel Goh}, \bibinfo{person}{Sandhini Agarwal}, \bibinfo{person}{Girish Sastry}, \bibinfo{person}{Amanda Askell}, \bibinfo{person}{Pamela Mishkin}, \bibinfo{person}{Jack Clark}, {et~al\mbox{.}}} \bibinfo{year}{2021}\natexlab{}.
\newblock \showarticletitle{Learning transferable visual models from natural language supervision}. In \bibinfo{booktitle}{\emph{International conference on machine learning}}. PMLR, \bibinfo{pages}{8748--8763}.
\newblock


\bibitem[Ramesh et~al\mbox{.}(2022)]%
        {dall-e2}
\bibfield{author}{\bibinfo{person}{Aditya Ramesh}, \bibinfo{person}{Prafulla Dhariwal}, \bibinfo{person}{Alex Nichol}, \bibinfo{person}{Casey Chu}, {and} \bibinfo{person}{Mark Chen}.} \bibinfo{year}{2022}\natexlab{}.
\newblock \showarticletitle{Hierarchical text-conditional image generation with clip latents}.
\newblock \bibinfo{journal}{\emph{arXiv preprint arXiv:2204.06125}} \bibinfo{volume}{1}, \bibinfo{number}{2} (\bibinfo{year}{2022}), \bibinfo{pages}{3}.
\newblock


\bibitem[Ramesh et~al\mbox{.}(2021)]%
        {dall-e}
\bibfield{author}{\bibinfo{person}{Aditya Ramesh}, \bibinfo{person}{Mikhail Pavlov}, \bibinfo{person}{Gabriel Goh}, \bibinfo{person}{Scott Gray}, \bibinfo{person}{Chelsea Voss}, \bibinfo{person}{Alec Radford}, \bibinfo{person}{Mark Chen}, {and} \bibinfo{person}{Ilya Sutskever}.} \bibinfo{year}{2021}\natexlab{}.
\newblock \showarticletitle{Zero-shot text-to-image generation}. In \bibinfo{booktitle}{\emph{International Conference on Machine Learning}}. PMLR, \bibinfo{pages}{8821--8831}.
\newblock


\bibitem[Rombach et~al\mbox{.}(2022)]%
        {stablediffusion}
\bibfield{author}{\bibinfo{person}{Robin Rombach}, \bibinfo{person}{Andreas Blattmann}, \bibinfo{person}{Dominik Lorenz}, \bibinfo{person}{Patrick Esser}, {and} \bibinfo{person}{Bj{\"o}rn Ommer}.} \bibinfo{year}{2022}\natexlab{}.
\newblock \showarticletitle{High-resolution image synthesis with latent diffusion models}. In \bibinfo{booktitle}{\emph{Proc. of the IEEE/CVF Conference on Computer Vision and Pattern Recognition}}. \bibinfo{pages}{10684--10695}.
\newblock


\bibitem[Ruiz et~al\mbox{.}(2023a)]%
        {dreambooth}
\bibfield{author}{\bibinfo{person}{Nataniel Ruiz}, \bibinfo{person}{Yuanzhen Li}, \bibinfo{person}{Varun Jampani}, \bibinfo{person}{Yael Pritch}, \bibinfo{person}{Michael Rubinstein}, {and} \bibinfo{person}{Kfir Aberman}.} \bibinfo{year}{2023}\natexlab{a}.
\newblock \showarticletitle{Dreambooth: Fine tuning text-to-image diffusion models for subject-driven generation}. In \bibinfo{booktitle}{\emph{Proc. of the IEEE/CVF Conference on Computer Vision and Pattern Recognition}}. \bibinfo{pages}{22500--22510}.
\newblock


\bibitem[Ruiz et~al\mbox{.}(2023b)]%
        {hyperdreambooth}
\bibfield{author}{\bibinfo{person}{Nataniel Ruiz}, \bibinfo{person}{Yuanzhen Li}, \bibinfo{person}{Varun Jampani}, \bibinfo{person}{Wei Wei}, \bibinfo{person}{Tingbo Hou}, \bibinfo{person}{Yael Pritch}, \bibinfo{person}{Neal Wadhwa}, \bibinfo{person}{Michael Rubinstein}, {and} \bibinfo{person}{Kfir Aberman}.} \bibinfo{year}{2023}\natexlab{b}.
\newblock \showarticletitle{Hyperdreambooth: Hypernetworks for fast personalization of text-to-image models}.
\newblock \bibinfo{journal}{\emph{arXiv preprint arXiv:2307.06949}} (\bibinfo{year}{2023}).
\newblock


\bibitem[Saharia et~al\mbox{.}(2022)]%
        {imagen}
\bibfield{author}{\bibinfo{person}{Chitwan Saharia}, \bibinfo{person}{William Chan}, \bibinfo{person}{Saurabh Saxena}, \bibinfo{person}{Lala Li}, \bibinfo{person}{Jay Whang}, \bibinfo{person}{Emily~L Denton}, \bibinfo{person}{Kamyar Ghasemipour}, \bibinfo{person}{Raphael Gontijo~Lopes}, \bibinfo{person}{Burcu Karagol~Ayan}, \bibinfo{person}{Tim Salimans}, {et~al\mbox{.}}} \bibinfo{year}{2022}\natexlab{}.
\newblock \showarticletitle{Photorealistic text-to-image diffusion models with deep language understanding}.
\newblock \bibinfo{journal}{\emph{Advances in Neural Information Processing Systems}}  \bibinfo{volume}{35} (\bibinfo{year}{2022}), \bibinfo{pages}{36479--36494}.
\newblock


\bibitem[Sohl-Dickstein et~al\mbox{.}(2015)]%
        {2015-diffusion}
\bibfield{author}{\bibinfo{person}{Jascha Sohl-Dickstein}, \bibinfo{person}{Eric~A. Weiss}, \bibinfo{person}{Niru Maheswaranathan}, {and} \bibinfo{person}{Surya Ganguli}.} \bibinfo{year}{2015}\natexlab{}.
\newblock \bibinfo{title}{Deep Unsupervised Learning using Nonequilibrium Thermodynamics}.
\newblock
\newblock
\showeprint[arxiv]{1503.03585}~[cs.LG]


\bibitem[Sohn et~al\mbox{.}(2023)]%
        {styledrop}
\bibfield{author}{\bibinfo{person}{Kihyuk Sohn}, \bibinfo{person}{Nataniel Ruiz}, \bibinfo{person}{Kimin Lee}, \bibinfo{person}{Daniel~Castro Chin}, \bibinfo{person}{Irina Blok}, \bibinfo{person}{Huiwen Chang}, \bibinfo{person}{Jarred Barber}, \bibinfo{person}{Lu Jiang}, \bibinfo{person}{Glenn Entis}, \bibinfo{person}{Yuanzhen Li}, \bibinfo{person}{Yuan Hao}, \bibinfo{person}{Irfan Essa}, \bibinfo{person}{Michael Rubinstein}, {and} \bibinfo{person}{Dilip Krishnan}.} \bibinfo{year}{2023}\natexlab{}.
\newblock \bibinfo{title}{StyleDrop: Text-to-Image Generation in Any Style}.
\newblock
\newblock
\showeprint[arxiv]{2306.00983}~[cs.CV]


\bibitem[Song et~al\mbox{.}(2022)]%
        {ddim}
\bibfield{author}{\bibinfo{person}{Jiaming Song}, \bibinfo{person}{Chenlin Meng}, {and} \bibinfo{person}{Stefano Ermon}.} \bibinfo{year}{2022}\natexlab{}.
\newblock \bibinfo{title}{Denoising Diffusion Implicit Models}.
\newblock
\newblock
\showeprint[arxiv]{2010.02502}~[cs.LG]


\bibitem[Wang et~al\mbox{.}(2023)]%
        {style-diffusion}
\bibfield{author}{\bibinfo{person}{Zhizhong Wang}, \bibinfo{person}{Lei Zhao}, {and} \bibinfo{person}{Wei Xing}.} \bibinfo{year}{2023}\natexlab{}.
\newblock \showarticletitle{StyleDiffusion: Controllable Disentangled Style Transfer via Diffusion Models}. In \bibinfo{booktitle}{\emph{Proceedings of the IEEE/CVF International Conference on Computer Vision}}. \bibinfo{pages}{7677--7689}.
\newblock


\bibitem[Yu et~al\mbox{.}(2022a)]%
        {vq-gan}
\bibfield{author}{\bibinfo{person}{Jiahui Yu}, \bibinfo{person}{Xin Li}, \bibinfo{person}{Jing~Yu Koh}, \bibinfo{person}{Han Zhang}, \bibinfo{person}{Ruoming Pang}, \bibinfo{person}{James Qin}, \bibinfo{person}{Alexander Ku}, \bibinfo{person}{Yuanzhong Xu}, \bibinfo{person}{Jason Baldridge}, {and} \bibinfo{person}{Yonghui Wu}.} \bibinfo{year}{2022}\natexlab{a}.
\newblock \bibinfo{title}{Vector-quantized Image Modeling with Improved VQGAN}.
\newblock
\newblock
\showeprint[arxiv]{2110.04627}~[cs.CV]


\bibitem[Yu et~al\mbox{.}(2022b)]%
        {parti}
\bibfield{author}{\bibinfo{person}{Jiahui Yu}, \bibinfo{person}{Yuanzhong Xu}, \bibinfo{person}{Jing~Yu Koh}, \bibinfo{person}{Thang Luong}, \bibinfo{person}{Gunjan Baid}, \bibinfo{person}{Zirui Wang}, \bibinfo{person}{Vijay Vasudevan}, \bibinfo{person}{Alexander Ku}, \bibinfo{person}{Yinfei Yang}, \bibinfo{person}{Burcu~Karagol Ayan}, {et~al\mbox{.}}} \bibinfo{year}{2022}\natexlab{b}.
\newblock \showarticletitle{Scaling autoregressive models for content-rich text-to-image generation}.
\newblock \bibinfo{journal}{\emph{arXiv preprint arXiv:2206.10789}} \bibinfo{volume}{2}, \bibinfo{number}{3} (\bibinfo{year}{2022}), \bibinfo{pages}{5}.
\newblock


\bibitem[Zhang and Agrawala(2023)]%
        {controlnet}
\bibfield{author}{\bibinfo{person}{Lvmin Zhang} {and} \bibinfo{person}{Maneesh Agrawala}.} \bibinfo{year}{2023}\natexlab{}.
\newblock \showarticletitle{Adding conditional control to text-to-image diffusion models}.
\newblock \bibinfo{journal}{\emph{arXiv preprint arXiv:2302.05543}} (\bibinfo{year}{2023}).
\newblock


\bibitem[Zhang et~al\mbox{.}(2023)]%
        {inversion}
\bibfield{author}{\bibinfo{person}{Yuxin Zhang}, \bibinfo{person}{Nisha Huang}, \bibinfo{person}{Fan Tang}, \bibinfo{person}{Haibin Huang}, \bibinfo{person}{Chongyang Ma}, \bibinfo{person}{Weiming Dong}, {and} \bibinfo{person}{Changsheng Xu}.} \bibinfo{year}{2023}\natexlab{}.
\newblock \showarticletitle{Inversion-based style transfer with diffusion models}. In \bibinfo{booktitle}{\emph{Proceedings of the IEEE/CVF Conference on Computer Vision and Pattern Recognition}}. \bibinfo{pages}{10146--10156}.
\newblock


\end{thebibliography}

\newpage
\appendix
\section{appendix}

\subsection{Training Step}

\begin{figure}[h!]
    \centering
    \includegraphics[width=.85\linewidth]{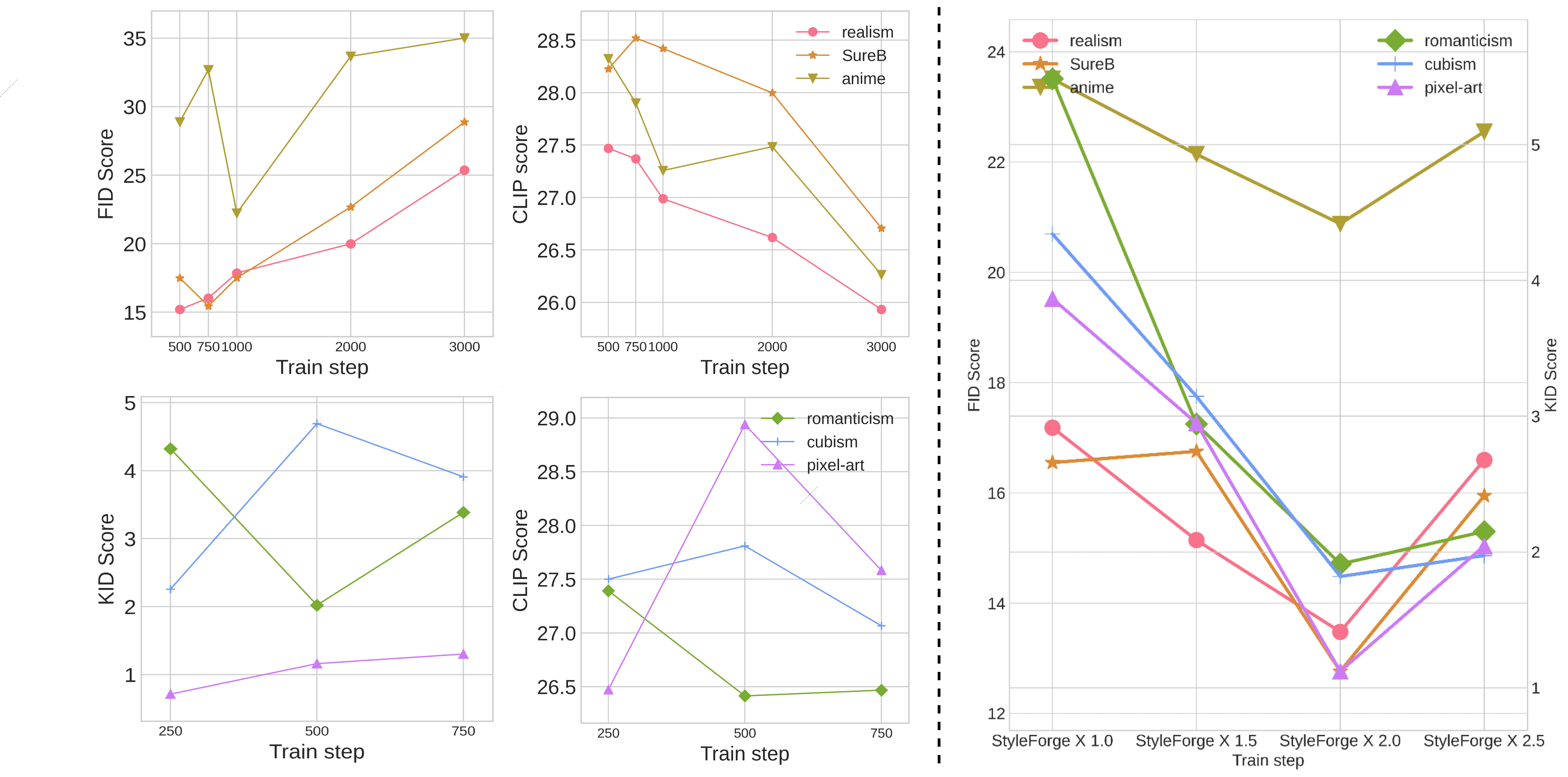}
    \caption{(left: Single-StyleForge) FID, KID ($\times 10^3)$, and CLIP scores of generated images as a function of processed fine-tuning steps for different target styles using only StyleRef images. The best FID scores are achieved at $500$, $750$, and $1000$ steps for \textsf{realism}, \textsf{SureB}, and \textsf{anime} styles, respectively. The best KID scores are achieved at $500$, $250$, and $250$ steps for \textsf{romanticism}, \textsf{cubism}, and \textsf{pixel-art} styles, respectively. 
    (right: Multi-StyleForge) The best KID/FID scores of Multi-StyleForge are achieved with doubling training steps of Single-StyleForge.
    }
    \label{fig:training_step}
    \vspace{-0.3cm}
\end{figure}

Fig.~\ref{fig:training_step} highlights the importance of adjusting training steps for each target style to optimize text-to-image synthesis. Personalizing the base model (SD v1.5), which predominantly generates realistic-like art images, to target styles such as \textsf{realism}, \textsf{romanticism}, \textsf{pixel-art} and \textsf{cubism} is relatively straightforward and achieves optimal FID/KID scores with fewer training steps. In contrast, adapting it to target styles less aligned with the base model, such as \textsf{SureB} and \textsf{anime}, requires extended iterations of $750$ and $1000$ training steps, respectively. 
Fewer training steps help mitigate overfitting, maintaining CLIP scores while preserving text-image alignment. Our approach, which focuses on personalizing styles, generally requires more training steps tha object-based personalization methods, e.g., DreamBooth~\cite{dreambooth}. However, it is observed that increasing training steps does not universally enhance style representation, as StyleRef images $\bfx$ encompass a limited style range and may neglect many visual attributes.


The training step influences both the denoising U-Net and the text encoder, which manages the text conditions in the diffusion process. Since multiple StyleRef prompts are used in Multi-StyleForge, more training steps are needed for the text encoder to effectively construct the latent space. This is illustrated in the right panel of Fig.~\ref{fig:training_step}. 
In the experiment, FID/KID scores and CLIP scores were compared across different training steps, starting with the same number of steps as Single-StyleForge and increasing by $2.5$ times. Using the same training steps as Single-StyleForge results in suboptimal FID/KID scores due to insufficient learning of multiple StyleRef prompts, leading to ineffective style personalization. Generally, doubling the training steps shows optimal performance, but further increases lead to overfitting, resulting in a decrease in overall evaluation metrics.


\subsection{Auxiliary image}

\begin{figure}[t!]
    \centering
    \includegraphics[width=0.5\linewidth]{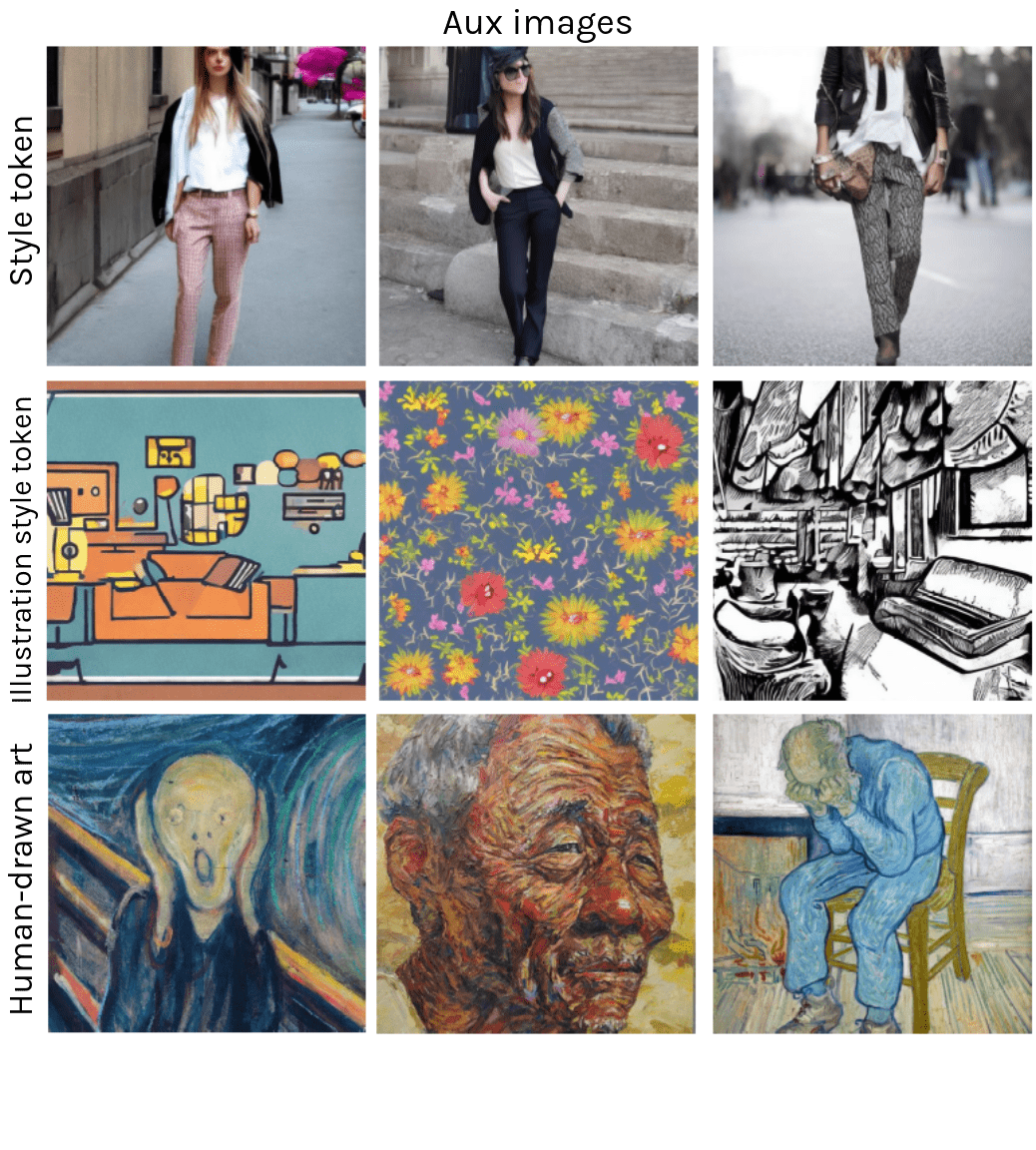}
    \caption{Results of different choices of Aux images $\bfx^\text{aux}$: (top) images crated from the frozen model using the \prompt{style} token; (middle) images created using the \prompt{illustration style} token; and (bottom) images drawn by a person. 
    }
    \label{fig:aux_image}
\end{figure}

Fig.~\ref{fig:aux_image} shows examples for each Aux image configuration. The first and second rows visualize the inherent information in the \prompt{style} and \prompt{illustration style} tokens of the pre-trained text-to-image diffusion model. The \prompt{style} token primarily contains semantic information related to fashion styles, while the \prompt{illustration style} token includes random artworks or textile patterns, showing significant differences compared to the human-drawn art images in the third row. This highlights the challenges of style personalization compared to object personalization. 
In object personalization, such as DreamBooth~\cite{dreambooth}, regularization images which play a similar role to auxiliary images provide clear and distinct support. For instance, if the meta-class name had been \prompt{dog}, the regularization images would directly reinforce the capability of producing the images of dog. However, when uncertain and less-specific auxiliary images (e.g., those associated with the \prompt{style} token in Fig.~\ref{fig:aux_image}) are used in training, unrefined information can interfere as negative auxiliary guidance in learning of StyleRef images.



\subsection{Qualitative comparison with baseline methods}
\begin{figure}[t!]
    \hspace*{-2cm}
    \centering
    \includegraphics[width=\textwidth]{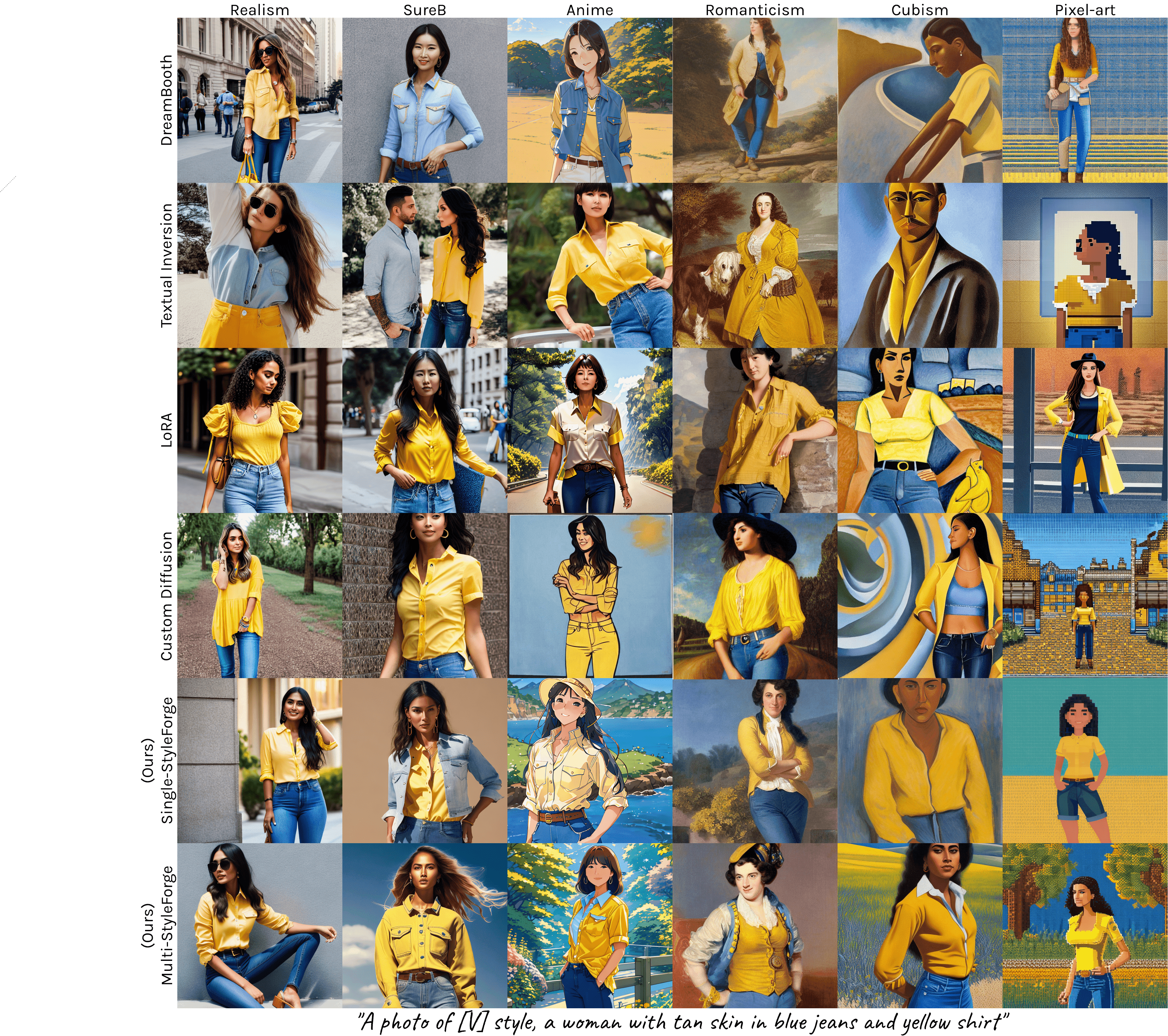}
    \caption{Comparison of our methods to existing personalization techniques. The images are guided by a prompt related to people: \prompt{a photo of [V] style, a woman with tan skin in blue jeans and yellow shirt}. Our models perform the desired synthesis by reflecting artistic styles and text, including detailed descriptions like \prompt{tan skin}.
    }
    \label{fig:comparison1}
    \vspace{-0.3cm}
\end{figure}

\begin{figure}[t!]
    \hspace*{-2cm}
    \centering
    \includegraphics[width=\textwidth]{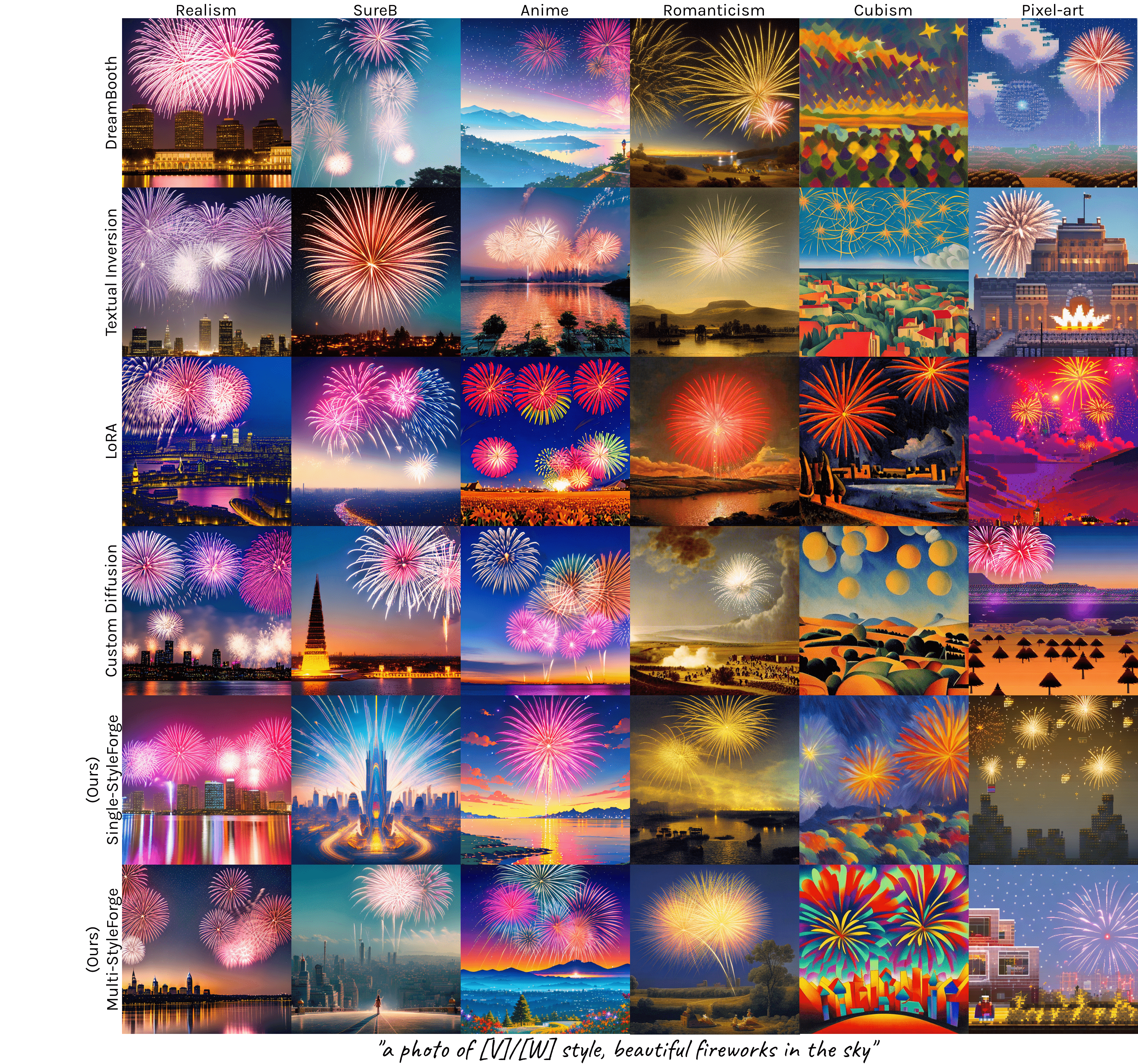}
    \caption{Comparison of our methods to existing personalization techniques. The images are guided by prompts related to the backgrounds.}
    \label{fig:comparison2}
\end{figure}

Fig.~\ref{fig:comparison1} and ~\ref{fig:comparison2}, along with Fig.~\ref{fig:comparison_total}, provide additional qualitative comparison results. Fig.~\ref{fig:comparison_total} illustrates images generated with prompts involving both people and backgrounds, Fig.~\ref{fig:comparison1} focuses on prompts solely related to people, and Fig.~\ref{fig:comparison2} focuses on backgrounds-related prompts. These comparisons allow us to assess the performance of our method in various scenarios, specifically verifying how effectively our model reflects both style and text in complex descriptions involving people and backgrounds. 

In Fig.~\ref{fig:comparison2}, we found that all methods including ours qualitatively reflect the art style and text well in background-related prompts. However, as specifically shown in Fig.~\ref{fig:comparison1}, baseline methods show a trade-off between capturing style and aligning text in content involving cognitive elements, such as people, where unnaturalness can cause significant errors. 
For example, in the case of Textual Inversion, it fails to capture the typical visual features of the \textsf{anime} style, e.g., vibrant colors, exaggerated facial expressions. 
Moreover, it is observed that most models including Single-StyleForge struggle to reflect some detailed text descriptions like \prompt{tan skin}. Conversely, Multi-StyleForge, which enhances text-image alignment, achieves superior visual reflection, effectively balancing style representation and detailed text descriptions.




\subsection{Applications}
\begin{figure*}[h!]
    \centerline{\includegraphics[width=\textwidth]{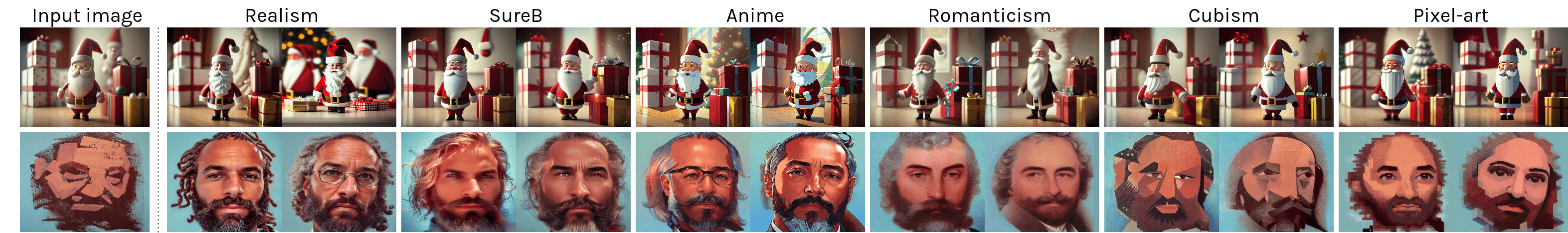}}
    \caption{Results of transforming input images (the leftmost) using Single-StyleForge. Output images were created with the prompts \prompt{a photo of [V] style, a Santa Clause} (first row), \prompt{a photo of [V], a man} (second row). Single-StyleForge synthesizes images that accurately reflect artistic styles, even when various forms of input images, including Santa Claus toys and watercolor brush paintings, are used. 
    }
    \label{fig:application}
    \vspace{-0.3cm}
\end{figure*}

Here, we demonstrate an interesting application of our approach. With the ability to incorporate specific styles into arbitrary images, our method leverages techniques from SDEdit~\cite{sdedit} to transform input images into specific artistic styles easily. The model iteratively denoises noisy input images (the leftmost images of Fig.~\ref{fig:application}) into the style distribution learned during training. Therefore, users only need to provide an image and simple text prompt, without requiring any particular artistic expertise or effort. Fig.~\ref{fig:application} showcases the results of styling input images with our Single-StyleForge method.


\end{document}